\pdfoutput=1

\documentclass[sigconf]{acmart} 

\usepackage{hyperref}
\hypersetup{
	colorlinks=true,
	linkcolor=blue,
	filecolor=red,      
	urlcolor=magenta,
	breaklinks=true,            
	draft=true,
}

\usepackage[utf8]{inputenc}

\setcitestyle{square}

\DeclareMathOperator*{\argmin}{argmin}

\def \OURS {\textit{Deferred Neural Rendering}}

\begin{document}
	\title{Deferred Neural Rendering:\\Image Synthesis using Neural Textures}
	
	\author{Justus Thies}
	\affiliation{%
		\institution{Technical University of Munich}}
	\email{justus.thies@tum.de}
	
	\author{Michael Zollh\"ofer}
	\affiliation{%
		\institution{Stanford University}}
	\email{zollhoefer@cs.stanford.edu}
	
	\author{Matthias Nie{\ss}ner}
	\affiliation{%
		\institution{Technical University of Munich}}
	\email{niessner@tum.de}
	
	\begin{abstract}
The modern computer graphics pipeline can synthesize images at remarkable visual quality; however, it requires well-defined, high-quality 3D content as input.
In this work, we explore the use of imperfect 3D content, for instance, obtained from photo-metric reconstructions with noisy and incomplete surface geometry, while still aiming to produce photo-realistic (re-)renderings.
To address this challenging problem, we introduce \OURS{}, a new paradigm for image synthesis that combines the traditional graphics pipeline with learnable components.
Specifically, we propose \textit{Neural Textures}, which are learned feature maps that are trained as part of the scene capture process.
Similar to traditional textures, neural textures are stored as maps on top of 3D mesh proxies; however, the high-dimensional feature maps contain significantly more information, which can be interpreted by our new deferred neural rendering pipeline.
Both neural textures and deferred neural renderer are trained end-to-end, enabling us to synthesize photo-realistic images even when the original 3D content was imperfect.
In contrast to traditional, black-box 2D generative neural networks, our 3D representation gives us explicit control over the generated output, and allows for a wide range of application domains.
For instance, we can synthesize temporally-consistent video re-renderings of recorded 3D scenes as our representation is inherently embedded in 3D space.
This way, neural textures can be utilized to coherently re-render or manipulate existing video content in both static and dynamic environments at real-time rates.
We show the effectiveness of our approach in several experiments on novel view synthesis, scene editing, and facial reenactment, and compare to state-of-the-art approaches that leverage the standard graphics pipeline as well as conventional generative neural networks.
	\end{abstract}
	
	\begin{teaserfigure}
		\centering
		\includegraphics[width=\textwidth]{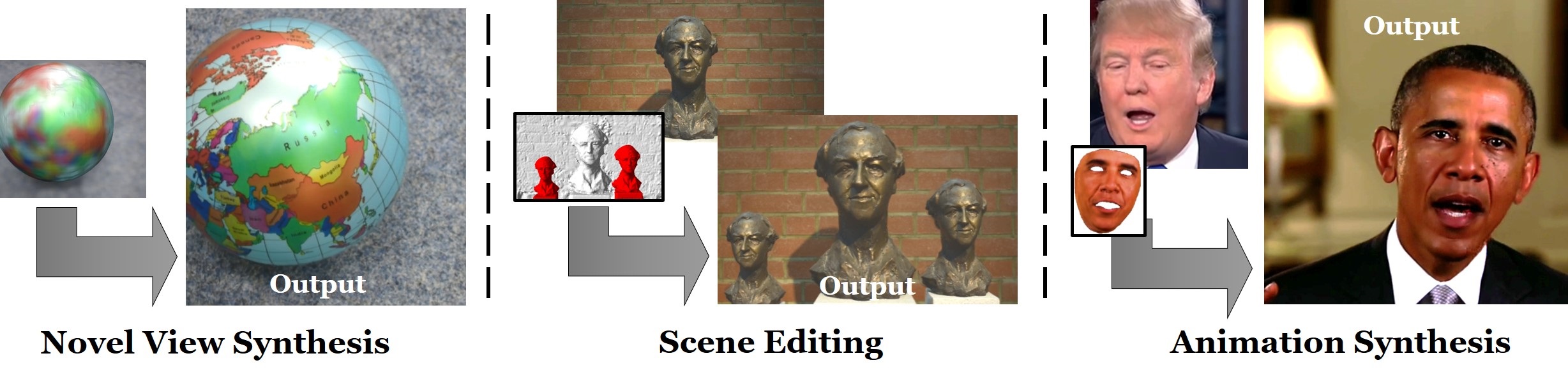}
		\caption{We present an image synthesis approach that learns object-specific neural textures which can be interpreted by a neural renderer. Our approach can be trained end-to-end with real data, allowing us to re-synthesize novel views of static objects, edit scenes, as well as re-render dynamic animated surfaces.}
		\label{fig:teaser}
		\vspace{0.25cm}
	\end{teaserfigure}


\setcopyright{none}
\settopmatter{printacmref=false} 
\renewcommand\footnotetextcopyrightpermission[1]{} 
\pagestyle{plain} 

\makeatletter
\def\runningfoot{\def\@runningfoot{}}
\def\firstfoot{\def\@firstfoot{}}
\makeatother

\maketitle

\section{Introduction}
\label{sec:intro}
The current computer graphics pipeline has evolved during the last decades, and is now able to achieve remarkable rendering results.
From a fixed function pipeline around the rasterization unit, the graphics pipeline has turned into a programmable rendering pipeline.
Based on this pipeline the Siggraph community has established rendering techniques that now achieve nearly photo-realistic imagery.
While the visuals are stunning, a major drawback of these classical approaches is the need of well-defined input data, including a precise definition of the surface geometry, the underlying material properties, and the scene illumination.
In movie or video game productions, this underlying 3D content is manually-created by skilled artists in countless working hours.
An alternative is to obtain 3D content from real-world scenes by using 3D reconstruction techniques.
However, given the inherent limitations of state-of-the-art 3D reconstruction approaches, such as noisy, oversmoothed geometry or occlusions, the obtained 3D content is imperfect.
From this captured content, it is nearly impossible to re-synthesize photo-realistic images with the existing computer graphics pipeline and rendering techniques.

In this work, we assume that captured 3D content will always suffer from reconstruction artifacts in one way or another.
Rather than aiming to fix the artifacts in the 3D content, we propose to change the paradigm of the rendering pipeline to cope with these imperfections.
To this end, we introduce {\em Deferred Neural Rendering} which makes a step towards a learnable rendering pipeline, combining learned {\em Neural Textures} with the traditional graphics pipeline.
Neural textures, similar to classical textures, are stored in 2D maps on top of a 3D mesh, and may be transformed along with the underlying 3D geometry.
However, the core idea behind neural textures is that they are composed of a set of optimal feature maps, rather than simple RGB values, that are learned during the scene capture process.
The rich signal stored in these high-dimensional neural textures encodes a high-level description of the surface appearance, and can be interpreted by our new deferred neural renderer.
Both the neural textures and the deferred neural renderer are trained in an end-to-end fashion, which enables us to achieve photo-realistic (re-)rendering results, even from imperfect geometry.
This learnable rendering pipeline enables a wide range of practical new application scenarios.
Our main motivation is to produce photo-realistic images from imperfect 3D reconstructions, for instance, obtained from a video sequence using multi-view stereo where the geometry typically is noisy, oversmoothed from strong regularization, or has holes.
Here, we show that we can synthesize novel view points of a scene while the geometry acts as a 3D proxy that forces generated images to be temporally-coherent with respect to the new camera view points.
However, the ability to explicitly control the 3D geometry not only allows synthesizing static view points, but also editing the underlying 3D content.
For instance, we can duplicate scene objects, along with their neural textures, and then coherently manipulate the re-renderings of the captured 3D environment.
We show several examples of such editing operations, e.g., copy or remove, in static 3D environments, which would have not been feasible with existing capture or editing techniques.
In addition, we demonstrate that we can edit dynamic scenes in a similar fashion.
Specifically, we show examples on human faces, where we first reconstruct the 3D face as well as the neural textures, and then modify the 3D facial animation and re-render a photo-realistic output.
This way, our unified neural rendering approach can easily achieve results that outperform the quality of existing facial re-enactment pipelines \cite{thies2016face}. 
In comparison to existing black-box, generative neural networks that are defined by a series of 2D convolutions, our approach is inherently embedded in 3D space.
As such, we implicitly obtain generated video output that is temporally coherent due to the underlying 3D embedding.
In addition, we have active control in manipulations, rather than making modifications to a semantically-uncorrelated random vector defining the latent space, as in 2D GAN approaches \cite{goodfellow2014generative}.
In summary, we combine the benefits of traditional graphics-based image synthesis with learnable components from the machine learning community.
This results in a novel paradigm of a learnable computer graphics pipeline with the following contributions:
\begin{itemize}
    \item \textit{Neural Rendering} for photo-realistic image synthesis based on imperfect commodity 3D reconstructions at real-time rates,
    \item \textit{Neural Textures} for novel view synthesis in static scenes and for editing dynamic objects,
    \item which is achieved by an end-to-end learned novel deferred neural rendering pipeline that combines insights from traditional graphics with learnable components.
\end{itemize}

\section{Related Work}
\label{sec:related}
\begin{figure*}[t!]
	\centering
	\includegraphics[width=\linewidth]{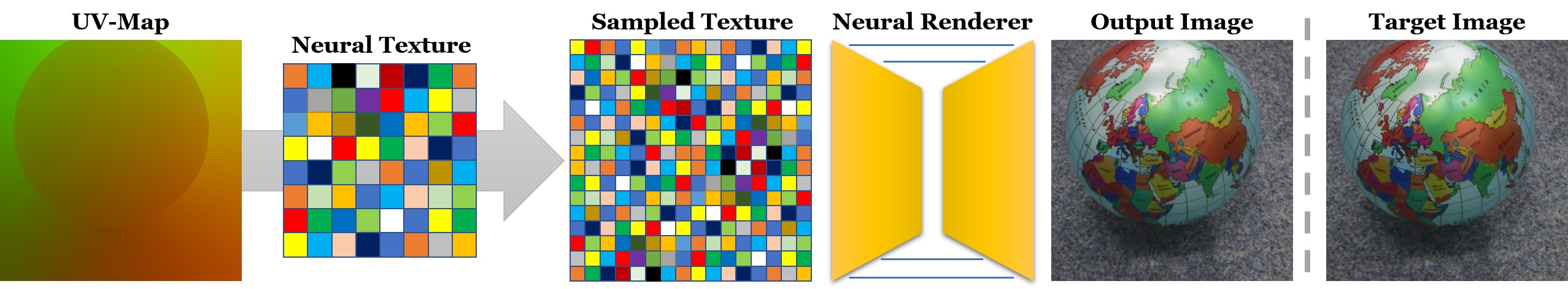}
	\caption
	{
        Overview of our neural rendering pipeline:
        Given an object with a valid uv-map parameterization and an associated \textit{Neural Texture} map as input, the standard graphics pipeline is used to render a view-dependent screen-space feature map.
        The screen space feature map is then converted to photo-realistic imagery based on a \textit{Deferred Neural Renderer}.
        Our approach is trained end-to-end to find the best renderer and texture map for a given task.
	}
	\label{fig:overview}
\end{figure*}

\OURS{} presents a new paradigm of image synthesis with learned neural textures and renderer.
Such learned computer graphics components can be useful for a variety of problems in computer graphics and computer vision.
In this work, we focus on novel view synthesis and synthesis of novel scene edits and animations, such as the animation of reconstructed human faces.

\subsection{Novel-view Synthesis from RGB-D Scans}
A traditional building block of novel-view synthesis approaches is to first obtain a digital representation of a real world scene.
In particular in the context of 3D reconstruction with commodity RGB-D sensors, researchers have made significant progress, enabling robust tracking \cite{izadi2011kinectfusion,newcombe2011kinectfusion,choi2015robust,whelan2016elasticfusion,dai2017bundlefusion} and large-scale 3D scene capture \cite{niessner2013real,chen2013scalable,zeng2013octree}.
Images from novel view points can be synthesized by rendering these reconstructions.
Given the inherent limitations in current state-of-the-art 3D reconstruction approaches, the obtained 3D content is imperfect, for instance, reconstructed geometry is noisy and/or oversmoothed, or has holes due to occlusion; this makes it nearly impossible to re-synthesize photo-realistic images.
A large body of work is also focused on the digitization of surface appearance; here, vertex colors can be estimated from the observations based on weighted averaging, but tracking drift and insufficient geometric resolution leads to blur.
Textures are an alternative that tackles the problem of missing spatial resolution; here, the geometry and color resolutions are decoupled, however, one must find a consistent uv-mapping.
One approach to compensate for camera drift and slightly wrong geometry is based on finding non-rigid warps \cite{zhou2014color,huang2017dlight}.
A major benefit of this line of work is that the reconstructed 3D content can be visualized by standard rendering techniques, and the approach directly generalizes to 4D capture as long as non-rigid surfaces can be reliably tracked \cite{newcombe2015dynamicfusion,innmann2016volumedeform,dou2016fusion4d}.
However, at the same time, imperfections in the reconstructions directly translate to visual artifacts in the re-rendering, which currently is the major hurdle towards making content creation from real-world scenes accessible.
\subsection{Image-based Rendering}
An alternative direction is to only utilize a very coarse geometry proxy and fill in the missing content based on high-resolution 2D textures \cite{huang2017dlight}.
Image-based rendering (IBR) pushes this to the limit, where the 3D geometry proxy is only used to select suitable views for cross-projection and view-dependent blending \cite{Buehler:2001,HeiglDAGM99,Carranza:2003,Zheng2009,hedman2016scalable}.
The advantage is that the visual quality of the re-rendered images does not exhibit the common artifacts caused by a low-resolution geometric representation.
However, many IBR approaches suffer from ghosting artifacts and problems at the occlusion boundaries.
These artifacts can be reduced using optical flow alignment \cite{EisemannFT,CasasRCTH15,Montage4D} or by different view-specific proxies \cite{Chaurasia:2013}.
An alternative is to directly model uncertainty \cite{Penner:2017}.
Our approach also changes the standard rendering pipeline to address the shortcomings of imperfect 3D surface geometry for rendering photo-realistic imagery.
However, in contrast to these approaches, \OURS{} is more flexible, since we learn so-called \emph{neural textures} that efficiently encode the appearance of an object in a normalized texture space.

\subsection{Light-field Rendering}
There exists a wide range of light-field rendering approaches~\cite{Gortler1996,Levoy1996,Magnor99adaptiveblock-based}.
In particular, our approach is closely related to the work on surface light fields.
Surface light fields store the direction-dependent radiance of every point on the surface, can be used for novel view synthesis, and are able to handle scenes with complex non-Lambertian surface properties.
Similar to our approach, a proxy geometry is required.
Lumitexels/Lumispheres are used to store the direction-dependent radiance samples at each surface point.
To store and query these lumitexels several different methods have been proposed \cite{Wood2000,Chen2002,Miandji2013}.
Our neural textures can be seen as a learned analog to these lumitexels, but instead of hand-crafted features, we employ end-to-end learning to find optimal features that can be interpreted by a neural network such that the original images are best reproduced.
Recently, Chen et al.~presented Deep Surface Light Fields \cite{Chen2018}, which reduces the required number of surface light field samples required for view interpolation using a neural network.
The used encoder-decoder network structure that estimates per-vertex colors is specially designed for the surface light field use case and learns to fill missing sample data across vertices.
Instead of predicting per vertex colors, we propose an end-to-end trained feature representation of scene appearance based on neural textures and a deferred neural renderer.
The deferred neural renderer is a convolutional network that takes the neighborhood of the rasterized surface points into account and, thus, is able to correct for reconstruction errors of the underlying geometry.
In contrast, the Deep Surface Light Fields approach needs high quality reconstructions, since a `slight misalignment can lead to strong artifacts such as ghosting and blurring'~\cite{Chen2018}.
In our results section, we demonstrate how our approach can handle a decreasing level of geometry quality and show a comparison to a per-pixel fully connected network.
\subsection{Image Synthesis using Neural Networks}
With the recent success of deep learning, neural networks can now also be utilized to synthesize artificial 2D imagery.
In particular, generative adversarial networks (GANs) \cite{goodfellow2014generative} and auto-regressive networks \cite{Oord:2016} achieve very impressive results in synthesizing individual images.
Pure synthesis approaches can be extended to a conditional setting, which is normally tackled with generator networks that follow a classical encoder-decoder architecture \cite{HintoS2006,jKingma2014}.
Conditional synthesis can be used to bridge the gap between two different domains, i.e., renderings of incomplete computer vision reconstructions and photo-realistic imagery.
Nowadays, conditional GANs (cGANs) are the de facto standard for conditional image synthesis \cite{RadfoMC2016,MirzaO2014,IsolaZZE2017}.
Recent approaches use generator networks based on a U-Net \cite{RonneFB2015} architecture, a convolutional encoder-decoder architecture with skip connections.
In both settings, high-resolution \cite{KarraALL2018,WangLZTKC2018} synthesis has been demonstrated.
While such generative approaches have shown impressive results for the synthesis of single, isolated images, synthesizing 3D and temporally-consistent imagery is an open problem.
One step in this direction is the vid2vid \cite{wang2018vid2vid} approach that employs a recurrent network for short-term temporal coherence.
Unfortunately, the produced results are not 3D consistent and lack photo-realism as well as long-term coherence.
The lack of 3D consistency is an artifact of trying to learn complex 3D relationships with a purely 2D image-to-image translation pipeline, i.e., in view-dependent screen space.
Rather than relying on a black-box neural network with a series of 2D convolutions, we propose \emph{neural textures}, which combine the knowledge of 3D transformations and perspective effects from the computer graphics pipeline with learnable rendering.
To this end, we efficiently encode the appearance of an object in normalized texture space of a 3D model, learned in an end-to-end fashion, and are hence able to combine the benefits from both graphics and machine learning.

\subsection{View Synthesis using Neural Networks}
Neural networks can also be directly applied to the task of view synthesis.
Novel views can be generated based on a large corpus of posed training images \cite{Flynn2016}, synthesized by learned warping \cite{zhou2016view, Park17}, or a layered scene representation \cite{lsiTulsiani18,Zhou:2018}.
Disentangled representations with respect to rotation or other semantic parameters can be learned using deep convolutional inverse graphics networks \cite{Kulkarni2015}.
Images can be synthesized based on low-dimensional feature vectors \cite{yan2016perspective,eslami2018neural} that obey geometric constraints \cite{worrall2017interpretable,cohen2014transformation}.
For example, latent variables based on Lie groups can be used to better deal with object rotation \cite{falorsi2018explorations}.
Recent learning-based IBR-approaches learn the blend function using a deep network \cite{hedman2018deep}, employ separate networks for color and disparity estimation \cite{LearningViewSynthesis}, explicitly learn view-dependent lighting and shading effects \cite{thies2018ignor}, or integrate features into a persistent Cartesian 3D grid \cite{Sitzmann:2018:DeepVoxels}.
While these approaches enforce geometric constraints, they yet have to demonstrate photo-realistic image-synthesis and do not directly generalize to dynamic scenes.
Our approach combines 3D knowledge in the form of neural textures with learnable rendering and gives us explicit control over the generated output allowing for a wide range of applications.

\section{Overview}
\label{sec:overview}
\begin{figure}
	\centering
	\includegraphics[width=\linewidth]{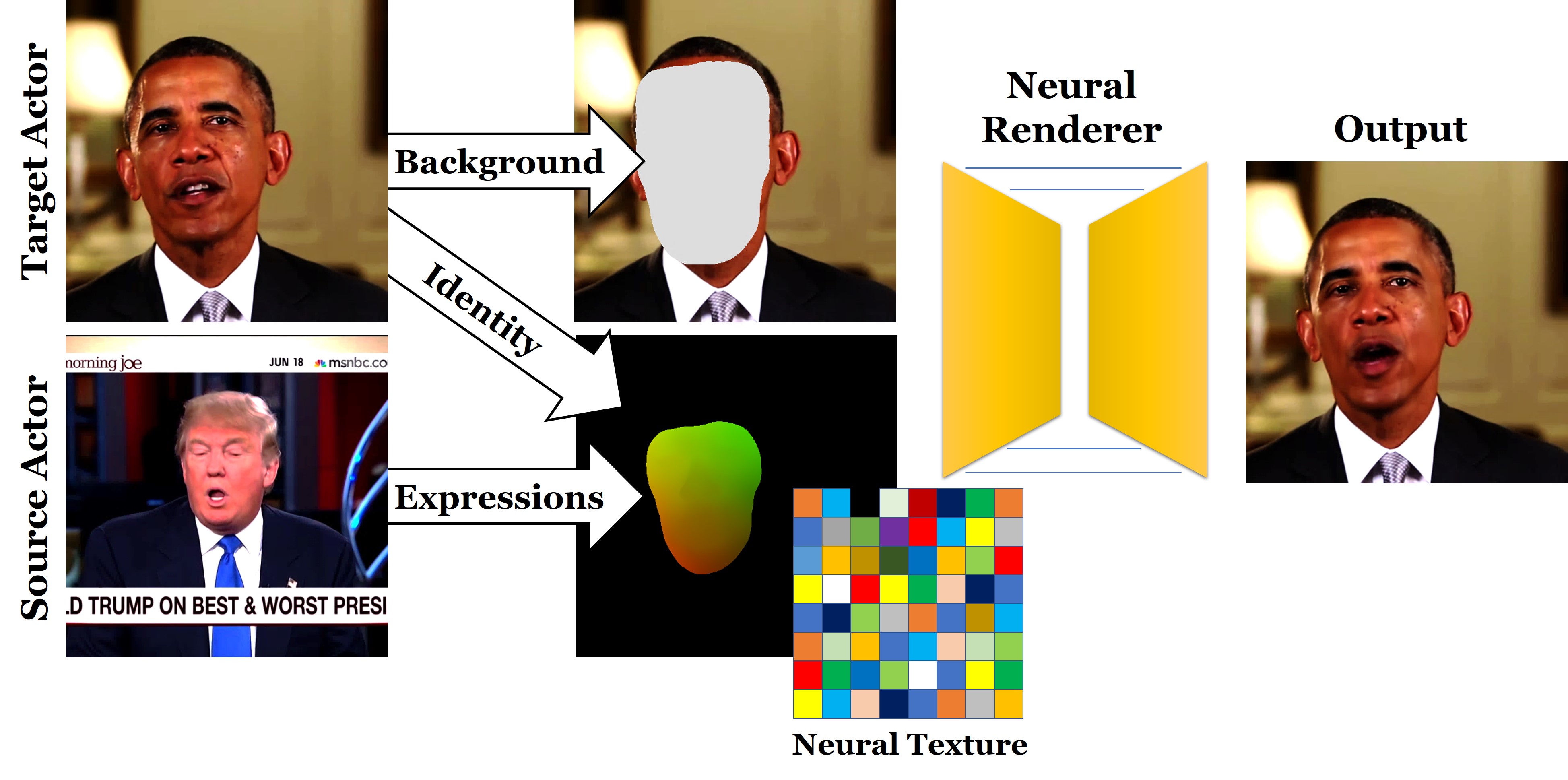}
	\caption
	{
        Overview of our reenactment synthesis pipeline.
        Using expression transfer, we generate an altered $uv$ map of the target actor matching  the expression of the source actor.
        This $uv$ map is used to sample from the neural texture of the target actor.
        In addition we provide a background image to the neural renderer, to output the final reenactment result.
	}
	\label{fig:overview_animation}
\end{figure}

The motivation for our work is to enable photo-realistic image synthesis based on imperfect commodity 3D reconstructions.
At the core of our approach are our \textit{Neural Textures} that are learned jointly with a \textit{Deferred Neural Renderer}.
Neural Textures are a new graphics primitive that can have arbitrary dimension and store a high-dimensional learned feature vector per texel.
Using the standard graphics pipeline, neural textures are sampled, resulting in a feature map in the target image space (see Fig.~\ref{fig:overview}).
Based on a trained \textit{Deferred Neural Renderer}, the sampled image space feature map is then interpreted.
The renderer outputs the final image that photo-realistically re-synthesizes the original object.

Neural Textures are the basis for a wide variety of applications ranging from novel-view synthesis to video editing.
Here, we concentrate on the use cases that are most relevant to computer graphics:
1) \textit{Neural Textures} can be used to texture a given mesh and, thus, can be easily integrated into the standard graphics pipeline.
In particular, for 3D scanned objects (e.g., via KinectFusion \cite{izadi2011kinectfusion,newcombe2011kinectfusion} or multi-view stereo reconstructions~\cite{colmap}), where additional ground truth color images are available, they enable learning photo-realistic synthesis of imagery from novel view points (see Fig.~\ref{fig:overview_animation}).
2) \textit{Neural Textures} can also be used to edit dynamic scenes in a similar fashion.
Specifically, we show reenactment examples for human faces (e.g., Face2Face \cite{thies2016face}), where we first reconstruct the 3D face as well as a neural texture, and then modify and realistically re-render the 3D facial animation.

In the following, we detail the stages of our deferred neural rendering pipeline, see Sec.~\ref{sec:main}.
Next, we show different use cases, including comparisons to approaches based on standard computer graphics algorithms, see Sec.~\ref{sec:results}.
Finally, we want to inspire the reader to further investigate the benefits of neural textures, see Sec.~\ref{sec:discussion}.

\section{Deferred Neural Rendering} 
\label{sec:main}

\OURS{} combines principles from the traditional graphics pipeline with learnable components.
The classical rendering pipeline consists of several stages that can potentially be made learnable.
In this work, we are focusing on \textit{Neural Textures} and \textit{Deferred Neural Rendering}, see Fig.~\ref{fig:overview}, which enable us to realize a variety of applications.
Lets consider the task of realistically re-rendering an object based on a noisy commodity 3D reconstruction.
Given a 3D reconstructed object with a valid $uv$-texture parameterization, an associated \textit{Neural Texture} map, and a target view as input, the standard graphics pipeline is used to render a view-dependent screen-space feature map.
This feature map is then converted to photo-realistic imagery based on a \textit{Deferred Neural Renderer}.
Our approach can be trained end-to-end to find the best renderer $\mathcal{R}$ and texture map $\mathbf{T}$ based on a training corpus of $N$ posed images $\{ \mathbf{I}_k, \mathbf{p}_k \}_{k=1}^{N} $.
Here, $\mathbf{I}_k$ is the $k$-th image of the training corpus and $\mathbf{p}_k$ are the corresponding camera parameters (intrinsic and extrinsic).
We phrase finding the best neural texture $\mathbf{T}^*$ and the best deferred neural renderer $\mathcal{R}^*$ for a specific task as a joint optimization problem over the complete training corpus:
\begin{equation}
	\mathbf{T}^*, \mathcal{R}^* = \argmin_{\mathbf{T}, \mathcal{R}}{ \sum_{k=1}^{N}
		{
			 \mathcal{L}( \mathbf{I}_k, \mathbf{p}_k | \mathbf{T}, \mathcal{R})
		}
	} \enspace{.}
\end{equation}
Here, $\mathcal{L}$ is a suitable training loss, i.e., a photometric re-rendering loss.
In the following, we describe all components in more detail.

\subsection{Neural Textures}
Texture maps are one of the key building blocks of modern computer graphics.
Typically, they contain appearance information, such as the albedo of an object, but they can also store custom attributes, such as high-frequency geometric detail in the form of normal or displacement maps.
These textures can be thought of as low-dimensional hand-crafted feature maps that are later on interpreted by programmed shader programs.
With this analogy in mind, \textit{Neural Textures} are an extension of traditional texture maps; instead of storing low-dimensional hand-crafted features, they store learned high-dimensional feature maps capable of storing significantly more information and can be interpreted by our new deferred neural rendering pipeline.
Instead of rebuilding the appearance of a specific object using hand-crafted features, we learn them based on a ground truth training corpus of images.
Both the neural textures and the deferred neural renderer are trained end-to-end, enabling photo-realistic image synthesis even if the original 3D content was imperfect.
In our experiments, we use $16$ feature channels.
It is possible to employ an intermediate re-rendering loss to enforce that the first $3$ feature channels represent the average color of the object, i.e., mimic a classical color texture map.

\subsection{Neural Texture Hierarchies}
Choosing the right texture resolution for a given scene and viewpoint is a challenging problem.
In particular, for complex scenes with high depth complexity there might not exist a single optimal choice.
Given a fixed texture resolution, if parts of the scene are far away from the viewer, the texture will be under-sampled, i.e., texture minification.
In contrast, if parts of the scene are close to the virtual camera, the texture is over-sampled, i.e., texture magnification.
Both minification and magnification might appear at the same time for different parts of the scene.
In classical computer graphics, Mipmaps are used to tackle this challenge.
Inspired by classical Mipmaps, we propose to employ \textit{Neural Texture Hierarchies} with $K$ levels.
We access the texture hierarchy by sampling values from all $K$ levels using normalized texture coordinates and bi-linear sampling.
The final color estimate is then obtained by adding all per-level sampling results (Laplacian Pyramid).
During training, our goal is to learn the best \textit{Neural Texture Hierarchy} that stores low frequency information on the coarse levels, while high frequency detail is represented on the finer levels.
We enforce this split during training based on a soft-constraint, i.e., we apply no regularization to the coarse levels and an increasing amount of $\ell_2$ regularization to the features channels of the finer levels.
\textit{Neural Texture Hierarchies} enable us to obtain higher quality results than with \textit{Neural Textures} alone; for an evaluation, see Fig.~\ref{fig:vase_sampling_issues}.
The improved quality is due to the fact that if only one high resolution neural texture would be used, which solves the minification problem, we run into overfitting problems during training due to texture magnification.
Overfitting the training data is problematic, since it leads to sampling issues during test time where unoptimized pixel values might be sampled.
Even though we do not explicitly specify a Mipmap interpolation scheme, we want to highlight that the network as well as the texture is trained end-to-end; thus, the 2D renderer has to learn a proper mapping of the Laplacian Pyramid along the entire training-set, including the compensation of aliasing effects in image space.

\subsection{Differentiable Sampling of Neural Textures}

One key property of traditional texture maps is that they can be sampled at arbitrary floating point image locations.
To this end, the returned color value is computed based on an interpolation scheme.
Our \textit{Neural Textures}, similar to the standard graphics pipeline, support bi-linear interpolation for sampling the stored high-dimensional feature maps in a differentiable manner.
This enables end-to-end training of our \textit{Neural Textures} together with the \textit{Deferred Neural Renderer}.
With the support for bi-linear sampling, we can use the standard graphics pipeline to rasterize the coarse proxy geometry and sample the neural texture.
This results in a view-dependent screen space feature map.
Note, during training we emulate the graphics pipeline, such that the forward and backward pass will exactly match the operations at test time.

\subsection{Deferred Neural Renderer}
\begin{figure}
	\centering
	\includegraphics[width=\linewidth]{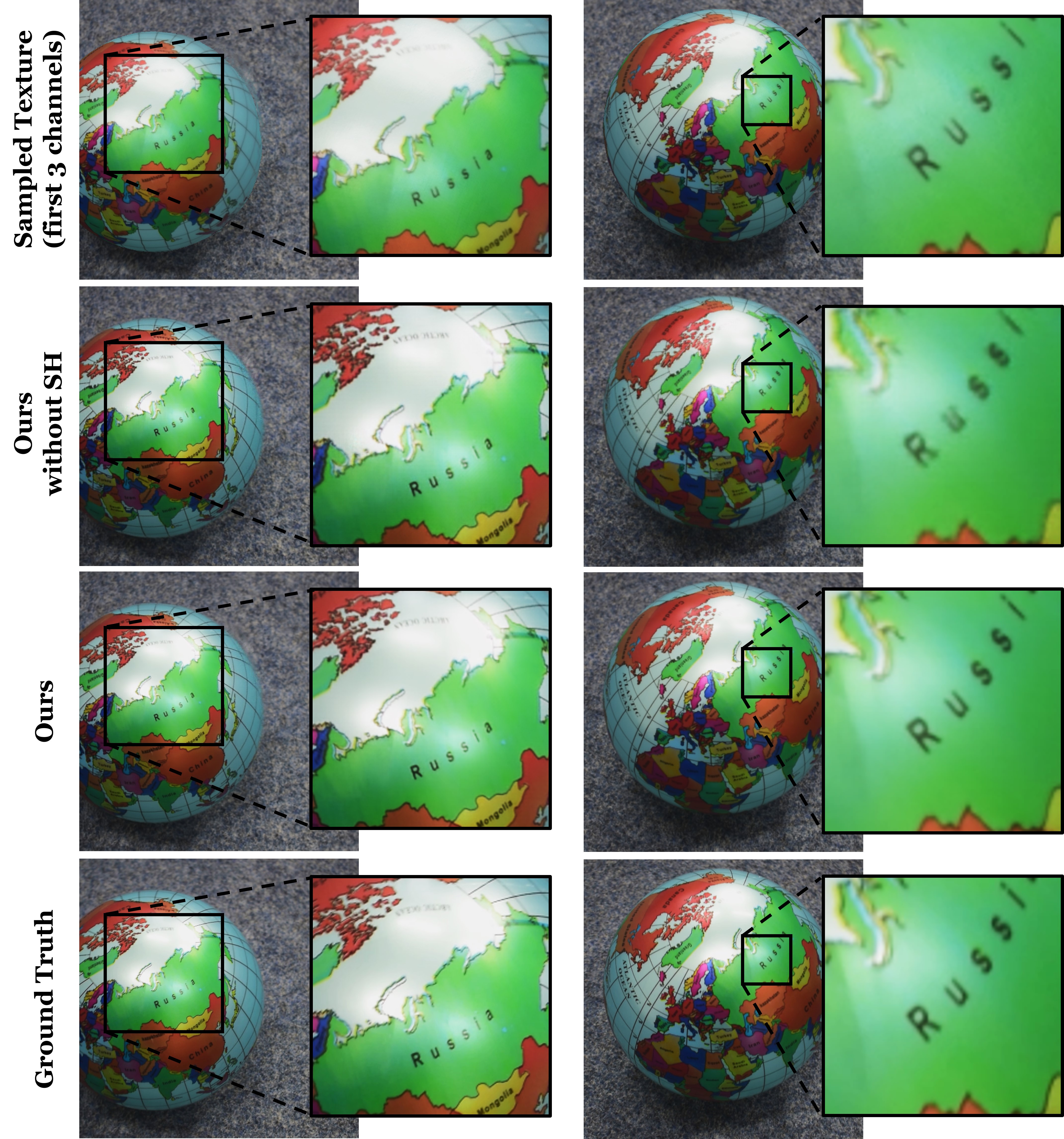}
	\caption
	{
        Novel View Synthesis: in contrast to a static mean texture, specular highlights are consistently reproduced with our approach.
        We also show the result without the spherical harmonics layer.
        As can be seen, the output is still photo-realistic, but it has a high MSE of $58.1$ compared to $48.2$ with our full pipeline (measured on a test sequence with $250$ images).
        The right column shows the differences in visual quality.
        With the spherical harmonics layer, the re-rendering is sharper and has a clearer specular highlight.
	}
	\label{fig:specular_highlights}
\end{figure}

The task of the \textit{Deferred Neural Renderer} is to form a photo-realistic image given a screen space feature map; this can be thought of as an analogy to classical deferred rendering.
We obtain the screen space feature map by rendering the coarse geometric object proxy, which is textured with the neural texture, from the ground truth viewpoint using the traditional rasterization pipeline.
Before training commences, we precompute the required texel-to-pixel mapping for each of the training images (referred to as $uv$ map in this paper).
At test time, the rasterizer of the graphics pipeline can be employed.
Based on this precomputed mapping, differentiable sampling can be used to obtain the screen space feature map via a lookup.
Our \textit{Deferred Neural Renderer} is based on recent advances in learning image-to-image mappings based on convolutional encoder-decoder network with skip-connection, similar to U-Net \cite{RonneFB2015} (see Appendix~\ref{sec:net_architecutre}).
Our network can have additional inputs such as a view-direction.
For view-dependent effects, we explicitly input the view-direction using the first $3$ bands of spherical harmonics resulting in $9$ feature maps.
Prior to the actual network, we multiply the sampled features with the evaluated spherical harmonic basis functions (feature channels $4-13$).
This allows us to rotate features with respect to the view-direction.
In Fig.~\ref{fig:specular_highlights}, we show the advantage of using such a spherical harmonics layer.
We train our neural texture representation end-to-end with the rendering network.

\begin{figure}
	\centering
	\includegraphics[width=\linewidth]{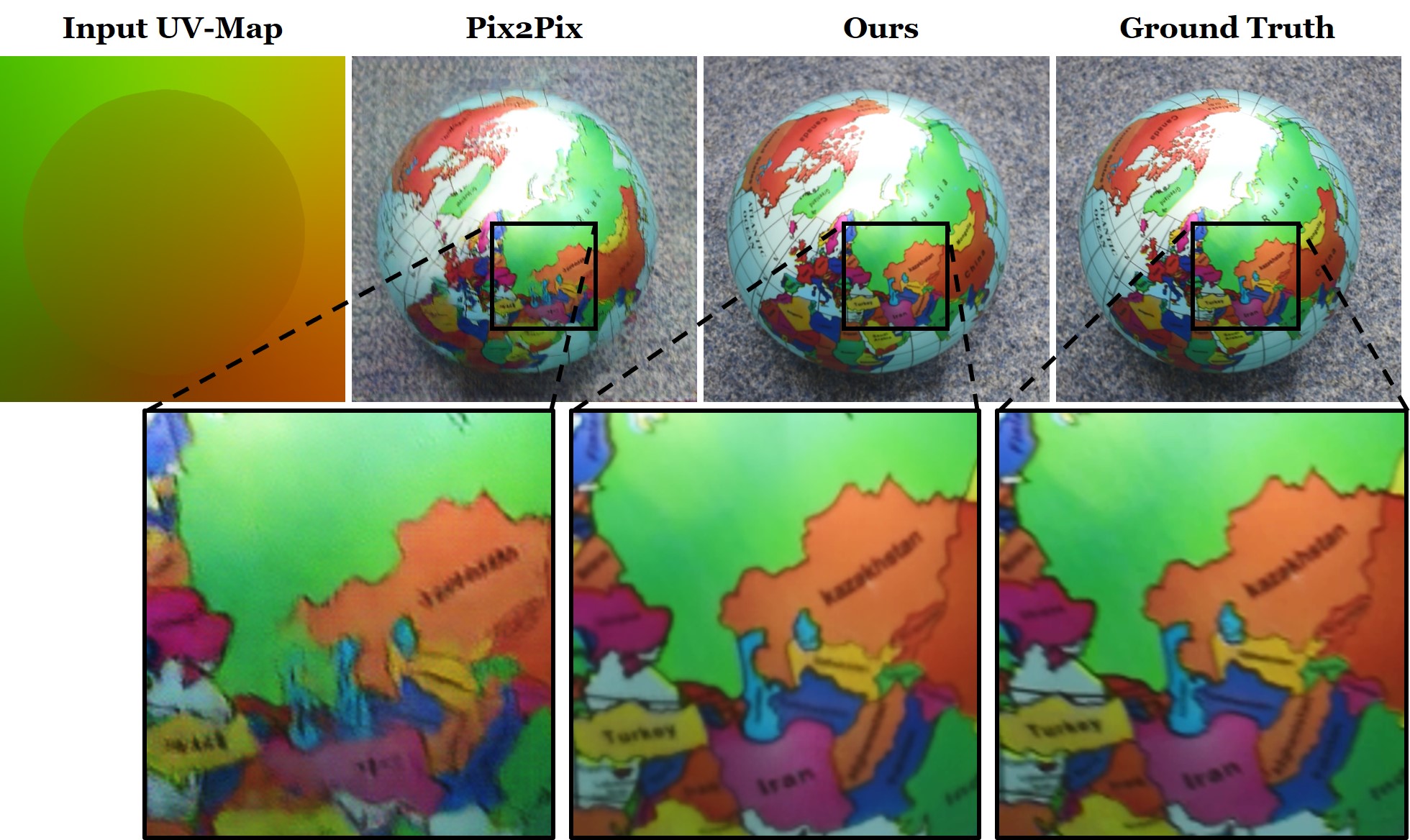}
	\caption
	{
        Comparison to the image-to-image translation approach Pix2Pix~\cite{pix2pix}. As can be seen, the novel views synthesized by our approach are higher quality, e.g., less blurry.
        Our results are close to the ground truth.
	}
	\label{fig:pix2pix}
\end{figure}

\subsection{Training}
We train our neural rendering approach end-to-end using stochastic gradient descent.
Since training images are static, we are able to precompute texture look-up maps which we will refer to as {\em uv-maps} in the following.
We build training pairs consisting of an $uv$-map and the corresponding ground truth color image.

In all experiments, we employ an $\ell_1$ photometric reproduction loss with respect to the ground truth imagery.
The $\ell_1$ loss is defined on a random crops (position and scale of the crops varies) of the full frame image.
Random cropping and scaling acts as a data augmentation strategy and leads to significantly better generalization to new views at test time.
We employ the Adam optimizer \cite{adam} built into PyTorch \cite{paszke2017automatic} for training.
The \textit{Deferred Neural Renderer} and the \textit{Neural Textures} are jointly trained using a learning rate of $0.001$ and default parameters $\beta_1=0.9$, $\beta_2=0.999$, $\epsilon=1\cdot e^{-8}$ for Adam.
We run approximately $50k$ steps of stochastic gradient descent.

\subsection{Training Data}
Our rendering pipeline is trained based on a video sequence that is also used to reconstruct the object.
The synthetic sequences consist of $1000$ random training views and a smooth trajectory of $1000$ different test views on the hemisphere.
The synthetic objects are provided by Artec3D\footnote{\url{https://www.artec3d.com/3d-models}}.
The average angular difference between the novel view direction and their nearest neighbor in the training set is $2.13^\circ$ (with a minimum of $0.04^\circ$ and a maximum of $6.60^\circ$).
In our ablation study w.r.t. the number of training images (see Fig.~\ref{fig:number_of_training_images}), the angular difference increases with a decreasing number of images: 500 images results in an average of $3.30^\circ$ ($0.11^\circ$ min / $9.76^\circ$ max), 250 images in $4.4^\circ$ ($0.11^\circ$/$12.67^\circ$) and 125 images in $6.27^\circ$ ($0.23^\circ$/$18.79^\circ$).
For real sequences, the training corpus size varies between $800$ and $1700$ frames depending on the sequence and the target application with similar angular differences from the test set to the training set (e.g.,~ Sequence~\ref{fig:coarse_geometry} with $800$ training images and a mean angular difference of $0.92^\circ$ ($0.04^\circ$/$3.48^\circ$), and Sequence~\ref{fig:classical_ibr} containing $1700$ training images with a mean angular difference of $0.85^\circ$ ($0.02^\circ$/$2.3^\circ$)).
Note that the angular differences in the viewing directions do not consider the positional changes in the cameras used for novel view synthesis.
For facial reenactment, a variety of facial expressions in the training video is necessary, otherwise it looses expressiveness.
In our experiments, we used $650$ training images for Macron, $2400$ for Obama, and $2400$ for Sequence~\ref{fig:face2face}.

\section{Results}
\label{sec:results}

The experiments on 2D neural textures are based on RGB input data.
We first reconstruct the geometry of the target object and estimate a texture parametrization.
The training data is created by re-rendering the $uv$-maps of the mesh that correspond to the observed images.
Using this training data, we optimize for the neural texture of the object, allowing us to re-render the object under novel views (see Sec.~\ref{sec:nvs_2d}) or animate the object (see Sec.~\ref{sec:animation}).
We used neural textures with a resolution of $512 \times 512$ with $4$ hierarchy level, containing $16$ features per texel and a U-Net with $5$ layers as a neural renderer.

\textit{The quality of our image synthesis approach can best be judged from the results shown in the supplemental video.}

\subsection{Novel View Point Synthesis}
\label{sec:nvs_2d}
For the novel view point synthesis of static objects, we require a 3D reconstruction and camera poses.
The training video sequence has been captured at a resolution of $1920 \times 1080$\,pixels at $30$\,Hz using a DSLR camera.
We obtain a coarse geometric proxy and the camera parameters (intrinsic and extrinsic) using the COLMAP \cite{colmap,colmapB} structure-from-motion approach.
To handle video sequences, which results in several hundred input images, we only use a subset of frames (every $25th$ frame) for the dense reconstruction.
The other frames are registered to this reconstruction.
The $uv$-parameterization is computed based on the Microsoft $uv$-atlas generator \footnote{\url{https://github.com/Microsoft/UVAtlas}}.

\begin{figure}
	\centering
	\includegraphics[width=\linewidth]{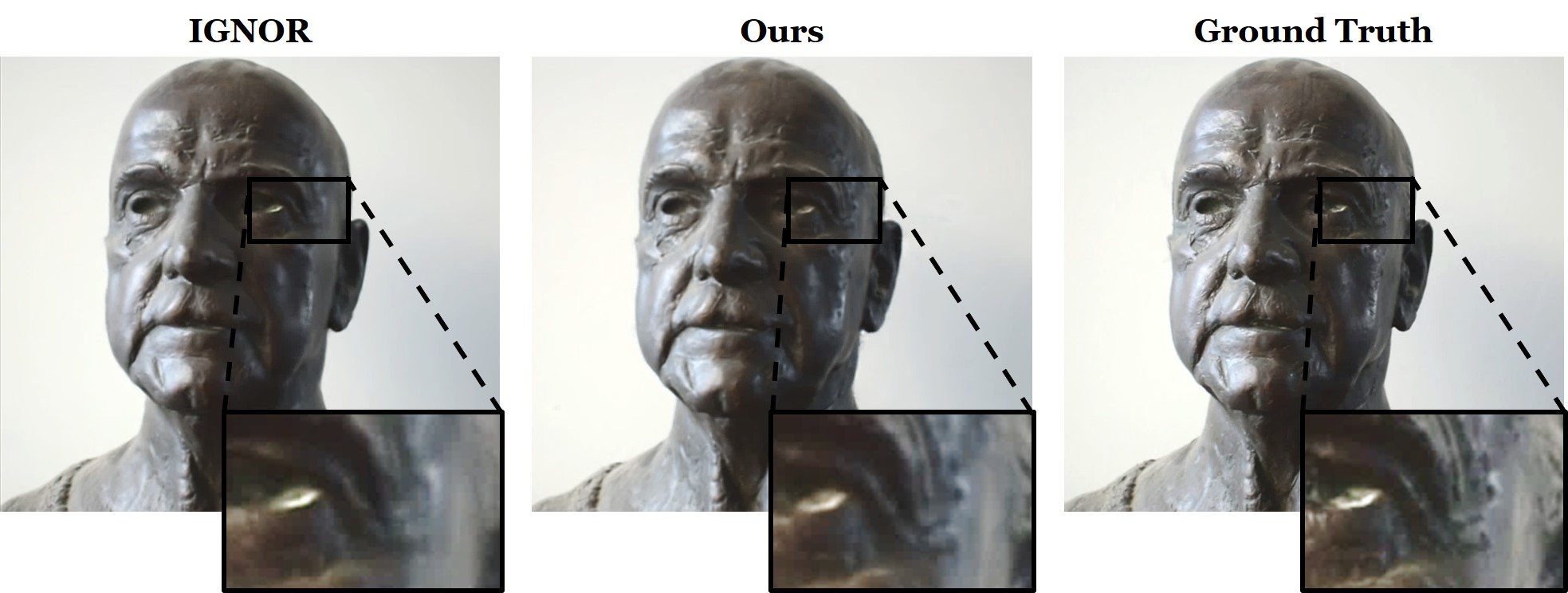}
	\caption
	{
        Comparison to the novel view point synthesis approach IGNOR~\cite{thies2018ignor}.
        Our approach better reproduces high frequency specular highlights.
	}
	\label{fig:comp_ignor}
\end{figure}

The learnable rendering pipeline allows us to re-render objects in a photo-realistic fashion.
We only consider one object during training time, which allows us to optimize for the object-specific texture and appearance.
Object-specific learning for novel view synthesis is known from recent publications~\cite{thies2018ignor,Sitzmann:2018:DeepVoxels}.
Fig.~\ref{fig:pix2pix} shows a novel view synthesis generated by our approach in comparison to an image-to-image translation network (Pix2Pix~\cite{pix2pix}) that directly predicts the output image based on the input $uv$-map.
Our approach outperforms Pix2Pix in terms of image quality with much sharper results.
Our results are also more temporally coherent (c.f. supplemental video).
In Fig.~\ref{fig:comp_ignor} we compare our approach to the image-guided neural object rendering approach (IGNOR) of \citet{thies2018ignor}.
In contrast to this image-guided rendering approach, our technique demonstrates a temporally more stable re-rendering of objects.
Our approach also does not need to store multiple of the original images for re-rendering, instead we store a single neural texture that enables us to synthesize the object under new views.

\begin{figure}
	\centering
	\includegraphics[width=\linewidth]{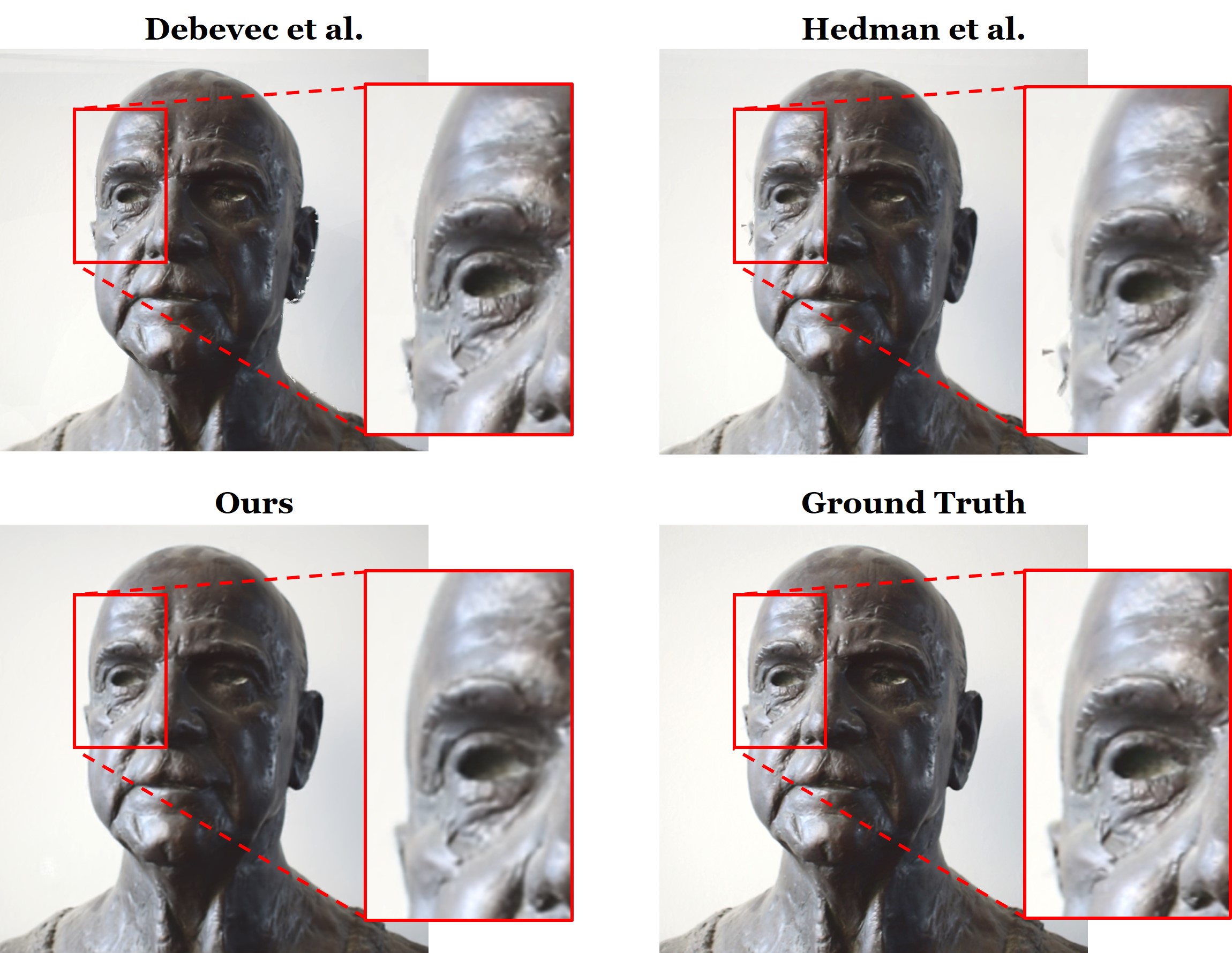}
	\caption
	{
        Comparison to Debevec et al.~\cite{Debevec1998} (based on $1686$ views) and Hedman et al.~\cite{hedman2016scalable} (based on the 99 views used for 3D reconstruction, since this approach requires the stereo reconstructed depth of the frames).
	}
	\label{fig:classical_ibr}
\end{figure}

In Fig.~\ref{fig:classical_ibr}, we also show comparisons to classical IBR methods.
As baseline we implemented the IBR method of Debevec et al.~\cite{Debevec1998}.
This approach directly reprojects the appearance of the captured views into the target view.
The original technique uses a per-triangle nearest neighbor view selection to reproject the appearance into the target view, instead, we do this view selection on a per-pixel basis, resulting in higher quality.
The nearest neighbors are selected among the entire training set (i.e., $1686$ views).
This per-pixel selection is far from real-time, but ensures optimal per-pixel results.
Note, that this IBR method heavily relies on the reconstructed geometry.
The MSE is $37.58$ compared to $20.65$ for our method. Especially, on occlusion boundaries the classical IBR method shows artifacts (see supplemental video).
The authors of Hedman et al.~\cite{hedman2016scalable} ran a comparison on the same sequence.
Besides the reconstructed model, they also use the reconstructed per frame depth.
Thus, their approach is limited to the frames used for the multi-view stereo reconstruction (in this sequence every $25th$ frame, resulting in a set of $99$ frames).
Their approach is able to generate high quality results with an MSE of $28.05$; improving the handling of occlusion boundaries in comparison to Debevec et al..
In comparison to both image-based rendering approaches, our technique does not require access to the training data during test time.
Only the texture ($512\times512\times16$) and the rendering network ($16$ million parameters) has to be provided which is magnitudes lower than storing hundreds of high resolution images.
In case of InsideOut~\cite{hedman2016scalable}, also the depth of each frame has to be stored.
Note that our approach needs to be trained on a specific sequence which is a drawback in comparison to the IBR methods. Training takes a similar amount of time as the stereo reconstruction and ensures high-quality synthesis of novel views.

%

\begin{figure}
	\centering
	\includegraphics[width=\linewidth]{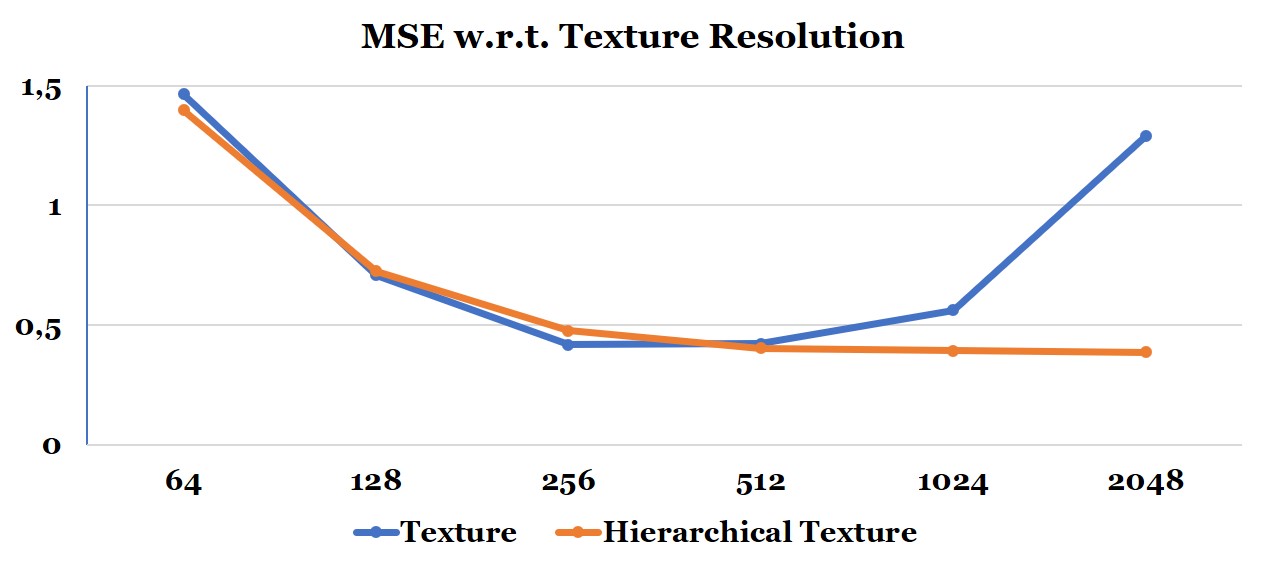}
	\caption
	{
        Influence of the neural texture resolution on the re-rendering quality.
        The graph is showing the MSE of the image synthesis vs.~the ground truth test data with respect to the neural texture resolution.
        In this experiment, a single texture without hierarchy has its sweet spot at a resolution of $256\times256$ achieving an MSE of $0.418$ with an increasing error for higher resolutions (at a resolution of $4096\times4096$ it reaches an MSE of $8.46$).
        In contrast, the hierarchical texture performs better on higher resolutions, i.e., an MSE of $0.38$ at a resolution of $2048\times2048$.
        The MSE is computed on color values in the range of $[0,255]$ using a test sequence of $1000$ frames, based on a synthetic rendering of a vase with Phong shading (see Fig.~\ref{fig:vase_sampling_issues}).
	}
	\label{fig:ablation_tex_dim}
\end{figure}

\begin{figure}
	\centering
	\includegraphics[width=\linewidth]{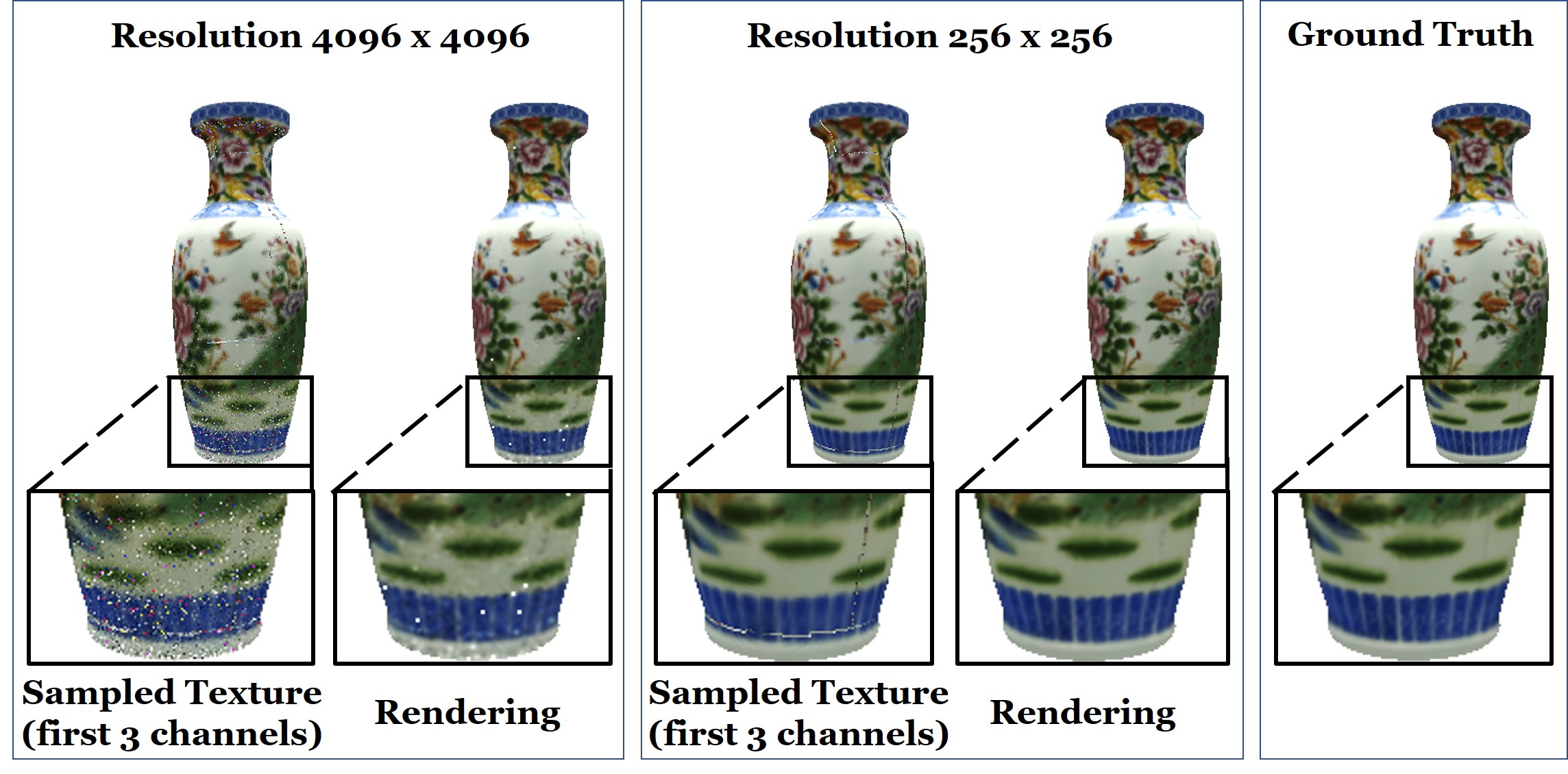}
	\caption
	{
        Influence of the neural texture resolution on re-rendering.
        Sample images of re-renderings using single neural textures.
        In addition to the re-renderings, we also show the first three channels of the sampled neural texture which are regularized to be the mean color.
	}
	\label{fig:vase_sampling_issues}
\end{figure}

\begin{figure}
	\centering
	\includegraphics[width=\linewidth]{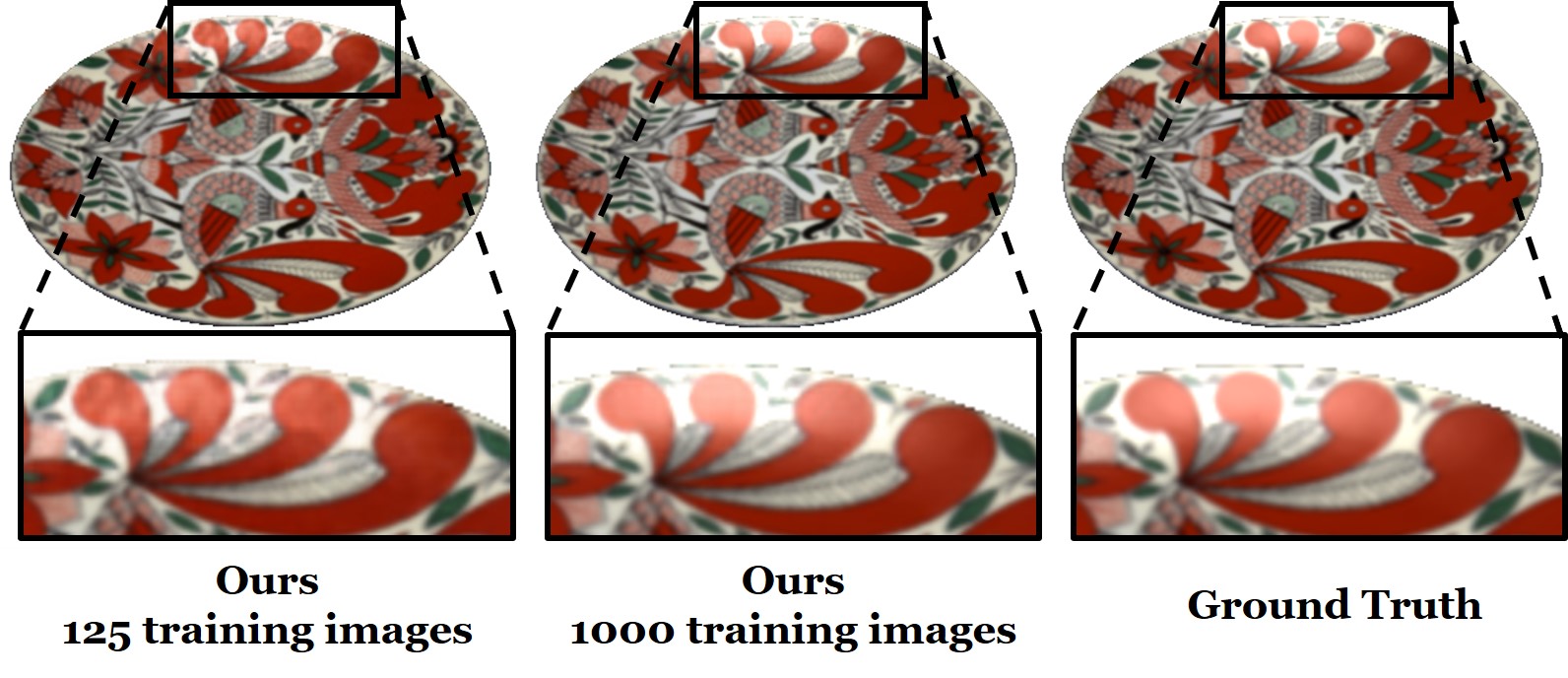}
	\caption
	{
        The number of images used for training influences the reconstruction ability.
        In particular, view-dependent effects such as specular highlights slowly degrade.
        Given the object above we achieve an MSE of $2.2$ for $1000$, $2.2$ for $500$, $6.5$ for $250$ and an MSE of $16.9$ for only $125$ images in the training set (assuming color values in the range of $[0,255]$ ).
	}
	\label{fig:number_of_training_images}
\end{figure}

\begin{figure}
	\centering
	\includegraphics[width=\linewidth]{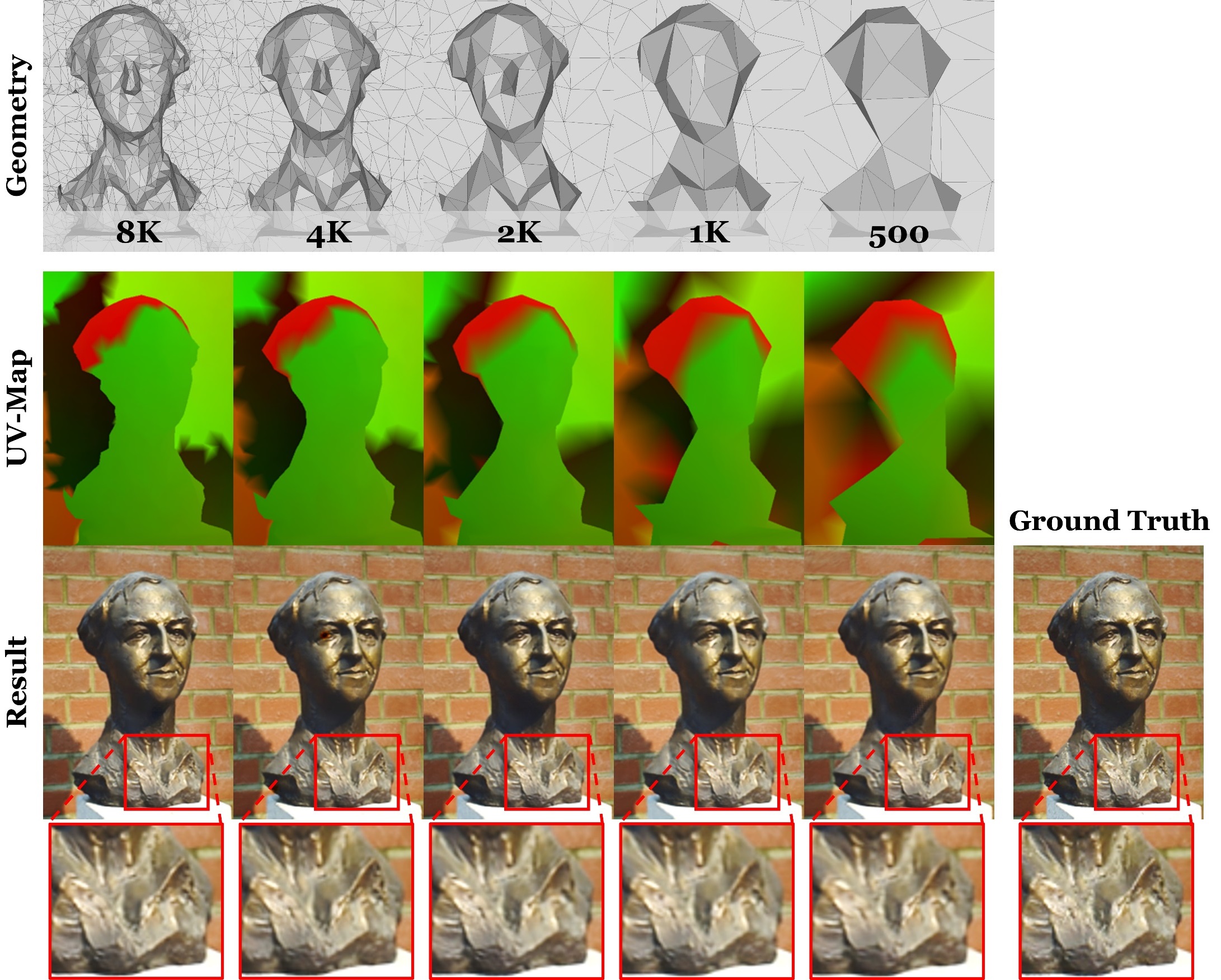}
	\caption
	{
	    Ablation study w.r.t. the resolution of the underlying geometry proxy.
	    Using quadric edge collapse we gradually reduce the number of triangles of the geometry from $8000$ to $500$.
	    Even with the lowest resolution, a photo-realistic image can be generated.
	    The MSE measured on the test sequence increases from $11.068$ (8K), $11.742$ (4K), $12.515$ (2K), $18.297$ (1K) to $18.395$ for a proxy mesh with $500$ triangles (MSE w.r.t. color channels in $[0,255]$).
	}
	\label{fig:coarse_geometry}
\end{figure}

%
%
Fig.~\ref{fig:ablation_tex_dim} and Fig.~\ref{fig:vase_sampling_issues} show a study on the influence of the resolution of the employed neural texture.
As can be seen, the best results for a single texture without hierarchy are obtained for a resolution of $256\times256$ achieving an MSE of $0.418$.
This sweet spot is due to tradeoffs in texture minification and magnification.
The hierarchical neural texture is able to further improve quality while increasing texture resolution, i.e., an MSE of $0.38$ at a resolution of $2048\times2048$.
Our approach is able to synthesize novel views based on a relatively small training set, but the quality of view-dependent effects, such as specular highlights, is slowly degrading with a decreasing number of training images (see Fig.~\ref{fig:number_of_training_images}).
In Fig.\ref{fig:coarse_geometry} we show an ablation study on the influence of the proxy geometry.
We gradually reduce the geometry resolution using quadric edge collapse.
As can be seen, our approach is also able to reproduce a reasonable output image given a very coarse mesh.

\begin{figure}
	\centering
	\includegraphics[width=\linewidth]{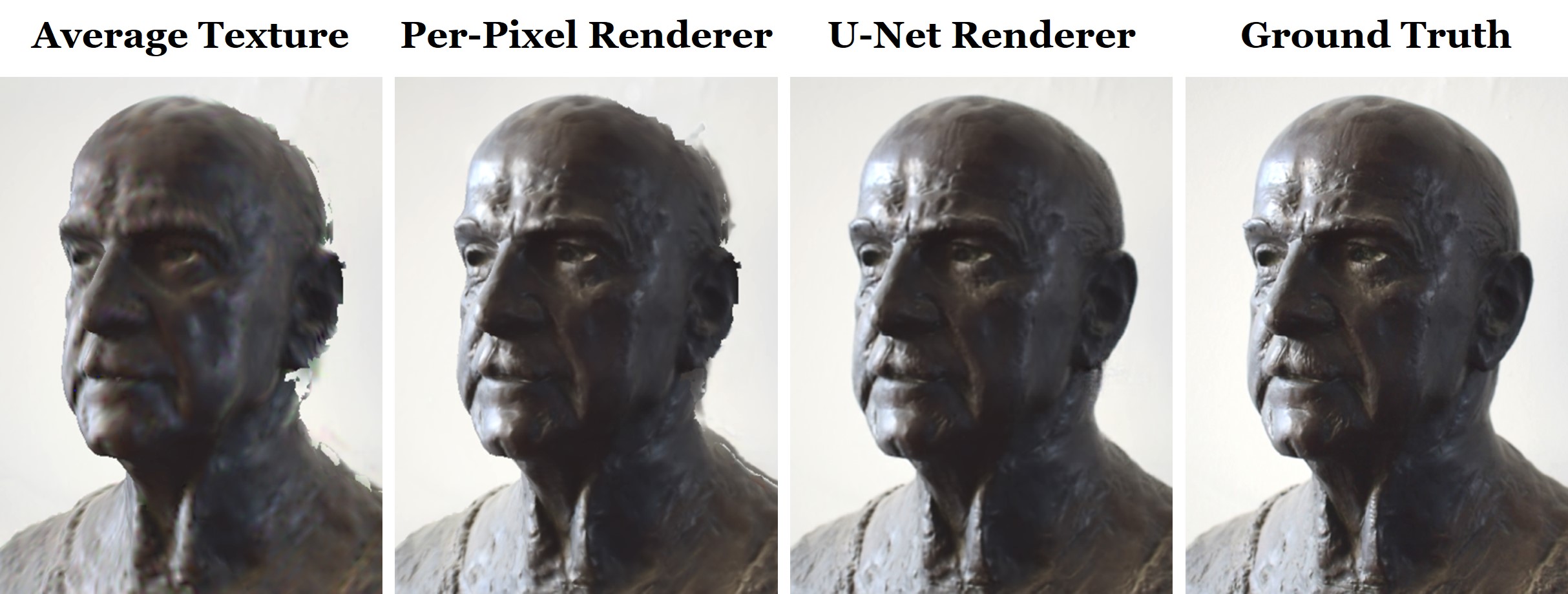}
	\caption
	{
        Here, we show a comparison of a U-Net renderer vs. a per-pixel fully connected render network.
        Since the U-Net renderer uses neighborhood information, it is able to correct the reconstruction errors of the underlying geometry.
        We also show the rendering based on a classical average texture.
	}
	\label{fig:u_net_vs_fc}
\end{figure}

To analyze the effect of the U-Net-like renderer, we also trained a per-pixel fully connected network that outputs the final image (see Fig.\ref{fig:u_net_vs_fc}).
Thus, this network does not leverage neighborhood information and is solely based on the per-pixel sampled texture values.
We use the same structure as our rendering network and replace all convolutions by $1 \times 1$ convolutions with stride $1$.
The idea of taking a per-pixel network is similar to the approach of Chen et al.~\cite{Chen2018} which estimates view-dependent per-vertex colors using a fully-connected network.
Since we are using a hand-held video sequence to reconstruct the object, the geometry contains errors.
As shown in our experiment, the neighborhood information is crucial to correct for these geometry errors. In contrast,  a per-pixel network leads to more blurry outputs (see supplemental video).
This observation is also mentioned in the paper of Chen et al.~\cite{Chen2018} and, thus, they have to take special care to reconstruct high quality geometry as input.

\begin{figure}
	\centering
	\includegraphics[width=\linewidth]{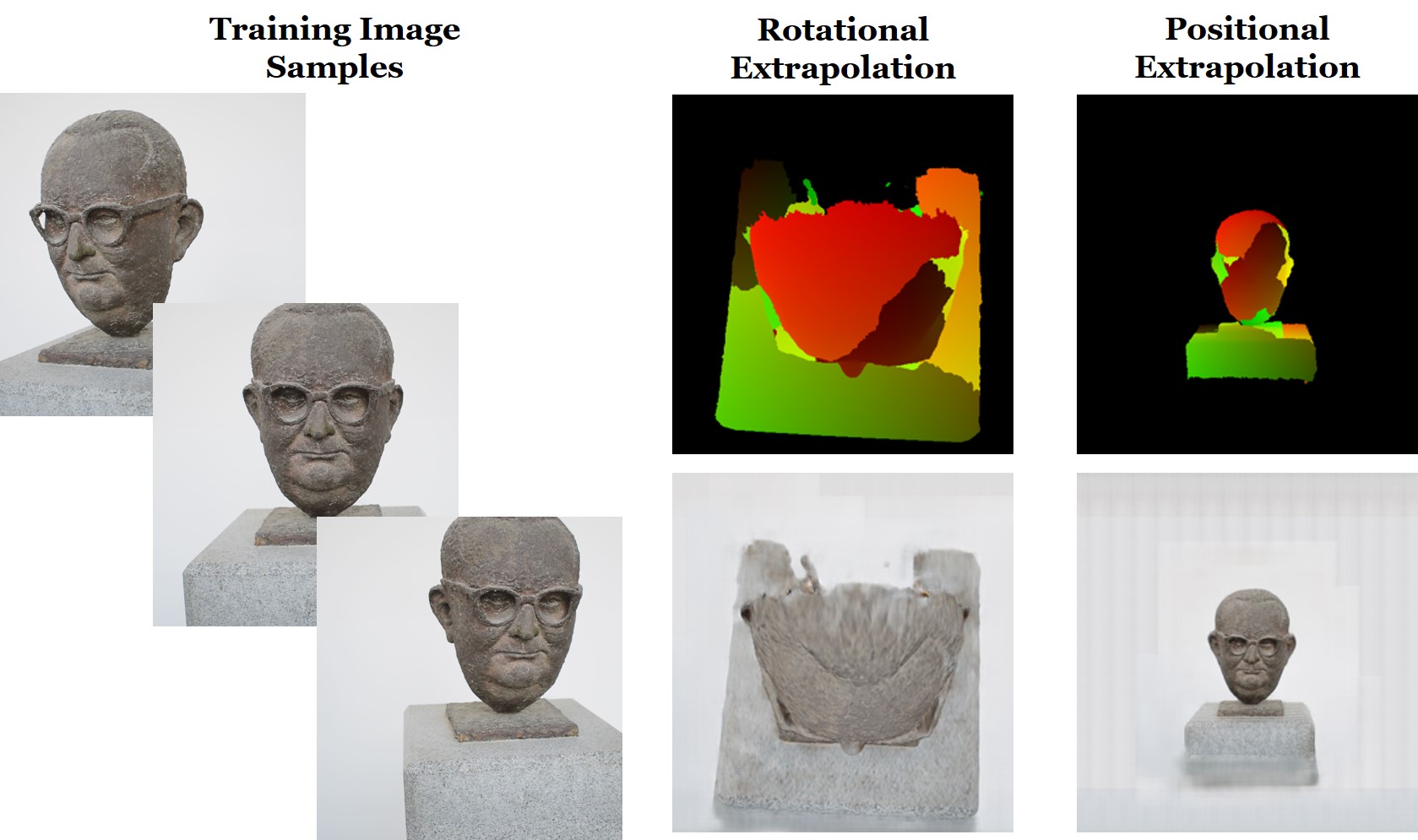}
	\caption
	{
	    Here, we show an extrapolation example, where we use a video of a bust captured from the front.
	    Both rotational and positional extrapolation leads to reasonable results.
	}
	\label{fig:extrapolation}
\end{figure}

Similar to other learning-based approaches, view extrapolation results in artifacts (see Fig.\ref{fig:extrapolation}).
As can be seen, our approach is still able to synthesize reasonable views of the object in the areas that where captured in the training video.

\begin{figure}[h!]
	\centering
	\includegraphics[width=\linewidth]{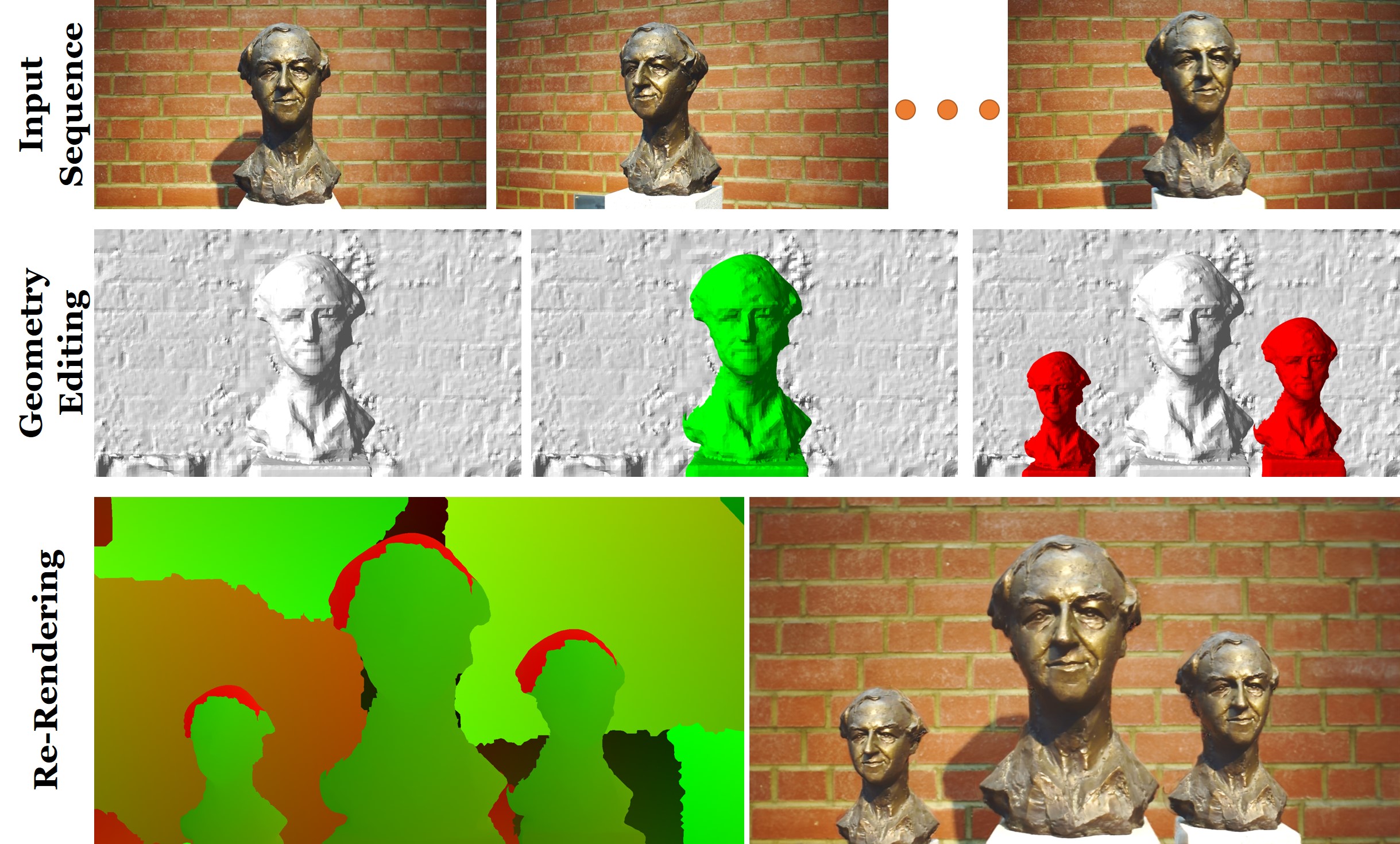}
	\caption
	{
        Overview of our scene editing pipeline.
        Given a video sequence, we estimate the geometry of the scene.
        We are then able to modify the geometry, e.g., by move and copy edits.
        During the edits, we keep track of the $uv$-parametrization of the vertices, which allows us to render a $uv$-map that is used as input to our neural renderer.
        The renderer and its corresponding texture is optimized using the original input images and the original geometry.
        Using the modified $uv$-map as input, we can apply the renderer to produce a photo-realistic output of the edited scene.
	}
	\label{fig:overview_edit}
\end{figure}

\begin{figure}[h!]
	\centering
	\includegraphics[width=\linewidth]{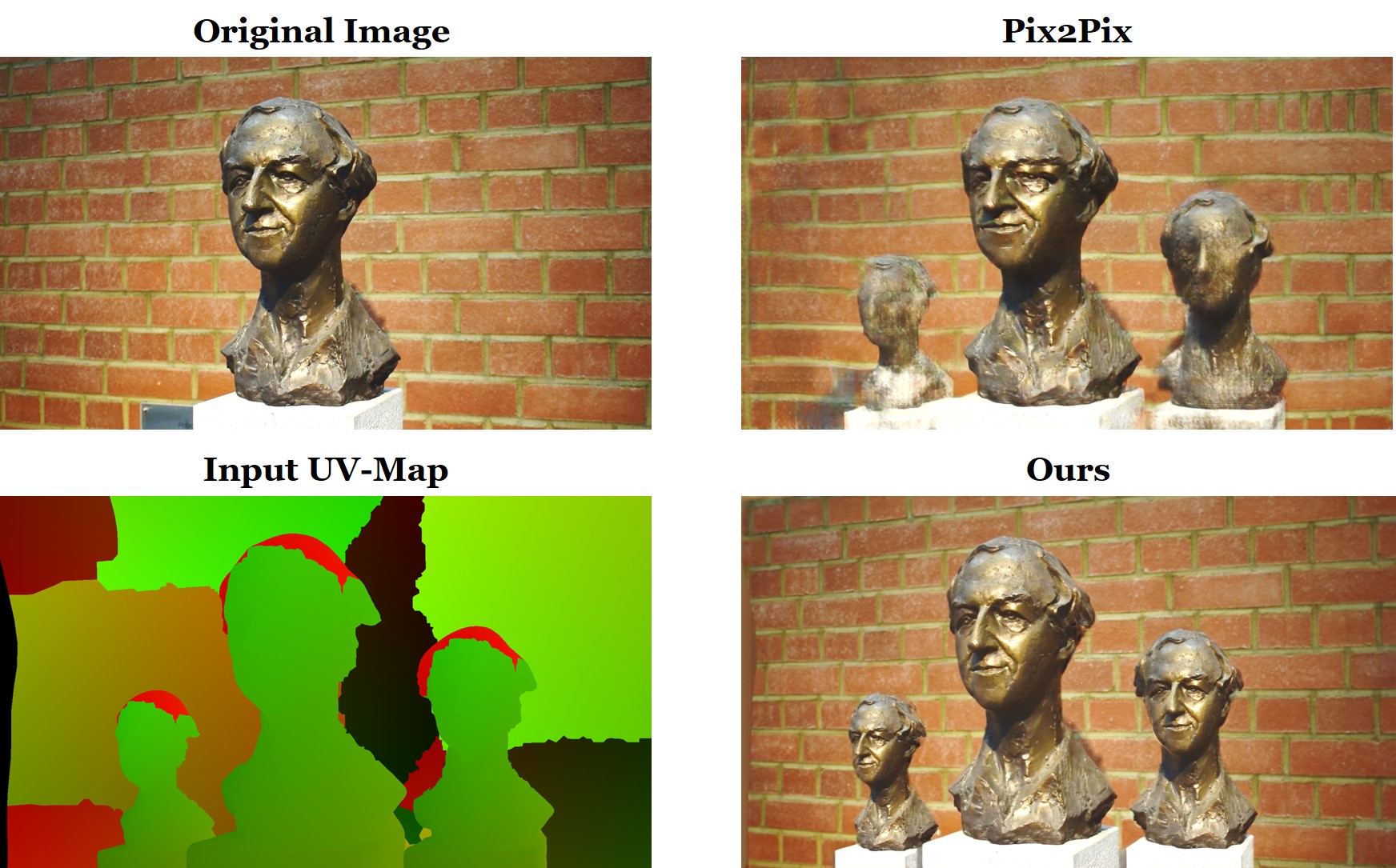}
	\caption
	{
        Editing comparison to Pix2Pix~\cite{pix2pix}.
        As can be seen, our approach better generalizes to scene edits.
	}
	\label{fig:scene_editing_pix2pix}
\end{figure}

\begin{figure}[h]
	\centering
	\includegraphics[width=\linewidth]{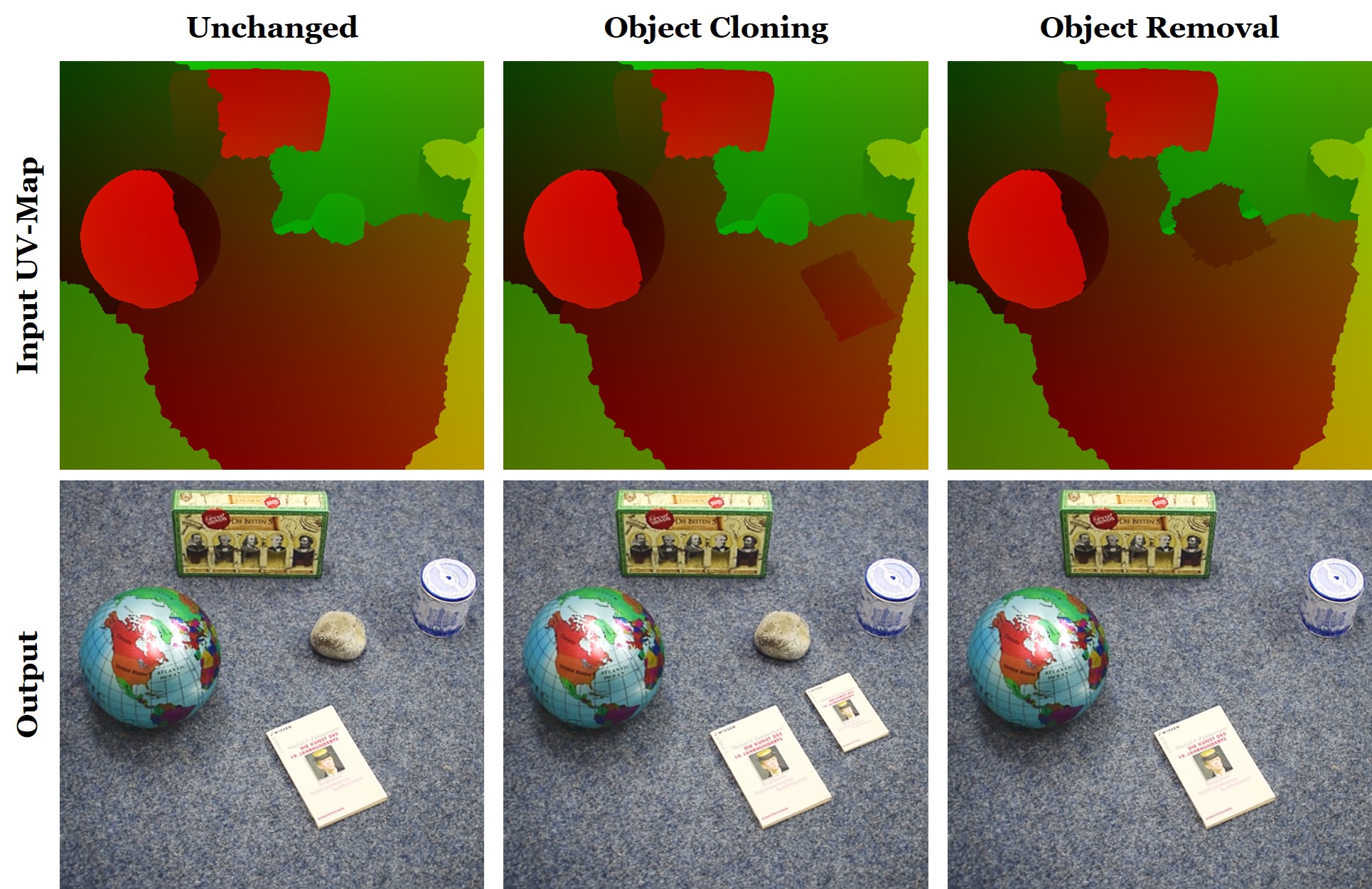}
	\caption
	{
        Editing a scene with multiple objects, including removal and cloning.
        Our approach obtains photo-realistic results.
	}
	\label{fig:scene_editing_removal}
\end{figure}

\paragraph{Scene Editing}
Using our rendering technique based on neural textures, we are able to apply edits to a scene (see Fig.~\ref{fig:overview_edit}).
Given a reconstruction of the scene, we can move objects around, remove, or duplicate them (see Fig.~\ref{fig:scene_editing_removal}).
In contrast to a standard image-to-image translation network, our method better generalizes to scene edits and generates temporally stable, high quality results (see Fig.~\ref{fig:scene_editing_pix2pix}).


\subsection{Animation Synthesis}
\label{sec:animation}

\begin{figure}[h]
	\centering
	\includegraphics[width=\linewidth]{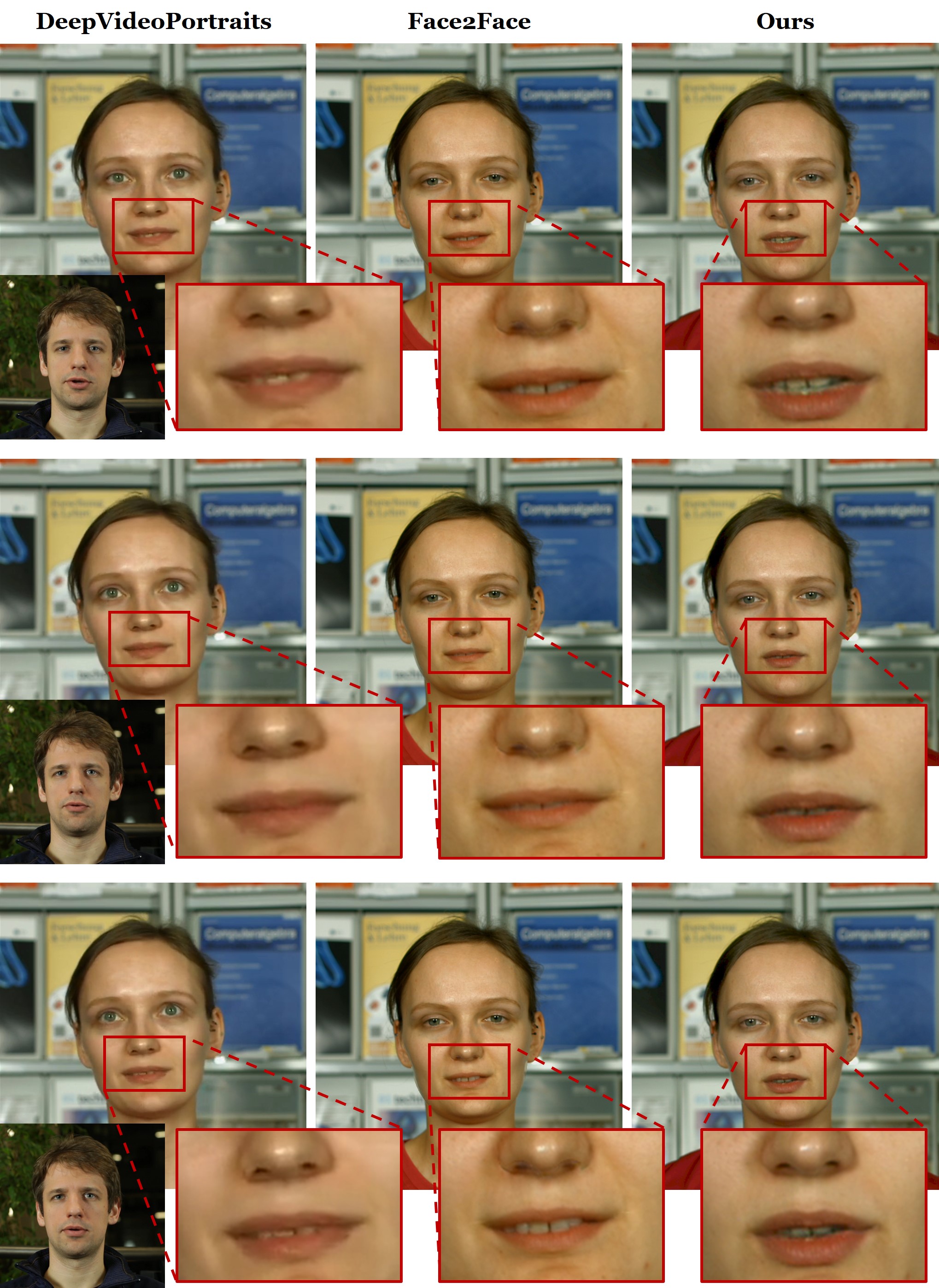}
	\caption
	{
        Comparison to the state-of-the-art facial reenactment approaches of \citet{kim2018DeepVideo} (DeepVideoPortraits) and \citet{thies2016face} (Face2Face).
        As can be seen, our learning-based approach better captures the person-specific idiosyncrasies of the target actor.
        The mouth interior has more variety, is much sharper, and has less stretching artifacts.
	}
	\label{fig:face2face}
\end{figure}

In this section, we show the capability to re-render dynamic objects, which allows us to implement applications such as facial reenactment (see Fig.~\ref{fig:overview_animation}).
To this end, we compare against the facial reenactment pipeline Face2Face~\cite{thies2016face}.
We leverage the 3D face reconstruction of Face2Face and the parametrization of the template mesh to generate the training data.
As a result of the training procedure, we optimize for a person-specific neural texture and a person-specific neural renderer.
We use the deformation transfer technique of Face2Face to re-render novel $uv$-maps of the target person corresponding to the expressions of a source actor which is the input to our technique.
Fig.~\ref{fig:face2face} shows the advantage of the trained neural texture.
It better captures the idiosyncrasies of the target actor, thus, enabling us to improve over state-of-the-art reenactment approaches that are either computer graphics based \citet{thies2016face} (Face2Face) or learned \citet{kim2018DeepVideo} (DeepVideoPortraits).
Note that we only have to train a single neural texture and renderer once for a target actor.
We do not have to store multiple textures for different expressions like in paGAN \cite{pagan}.
Reenactment can be done with any new source actor.
Even real-time reenactment is possible since the rendering only takes $\approx 4$ms (mean evaluation time for the test sequence in Fig.~\ref{fig:face2face} on a Nvidia 1080Ti) in comparison to $10$ms for the rendering process of Face2Face.

While our approach is able to generate photo-realistic reenactment results, it raises ethical concerns.
Our digital society is strongly relying on the authenticity of images and video footage.
On one hand researchers try to close the gap between computer generated images and real images to create fantastic looking movies, games and virtual appearances.
On the other hand, these techniques can be used to manipulate or to create fake images for evil purposes.
Image synthesis approaches, especially the ones related to humans, are therefore of special interest to the digital media forensics community.
Recently, techniques have been proposed that employ deep neural networks to detect such generated images~\cite{roessler2018faceforensics,roessler2019faceforensics++}.
The drawback of these methods is that they rely on a huge training set.
Transferability to other synthesis approaches is very limited, but is getting into the focus of researchers~\cite{cozzolino2018forensictransfer}.

\section{Conclusion}
\label{sec:discussion}
In this work, we have presented neural textures, and demonstrated their applicability in a large variety of applications: novel view synthesis, scene editing, and facial animation editing.
Interestingly, our rendering is also significantly faster than traditional reenactment, requiring only a few milliseconds for high-resolution output. 
However, we believe this is only a stepping stone to much wider range of applications where we use imperfect 3D content in order to generate photo-realistic output.
Our approach relies on a geometry proxy that has to be reconstructed in a preprocessing step.
If the geometry is too coarse, the result quality gracefully degrades, i.e., the re-rendered images get more blurry.
As the appearance of each object is unique, we must train the neural texture for every new object.
However, we believe that neural texture estimation based on a few images in conjunction with a generalized neural renderer is a very promising direction for future work.
Preliminary results on synthetic data suggest that our neural renderer can be generalized; currently, on real data, the size of our dataset (<10 objects) is limiting, but we believe that this generalization has potential for many new applications; e.g., transfer learning, where a renderer can be exchanged by another renderer that, for example, renders segmentations or the like.
The idea of neural textures is also not only bound to the two-dimensional setting that is discussed in our work.
It would be straightforward to extend the idea to higher dimensions or other data structures (e.g.,~a volumetric grid).
Beyond finding the right scene representation, there are also many other problems that can be tackled, such as disentangled control over the scene illumination and surface reflectance.
In our work, we assume static illumination and, thus, we are not able to relight the scene.
In addition, we believe there is a unique opportunity to revisit other components and algorithms of the traditional rendering pipeline, such as novel shading and lighting methods, or even use a full differentiable path tracing framework~\cite{li2018differentiable}.
In conclusion, we see a whole new field of novel computer graphic pipeline elements that can be learned in order to handle imperfect data captured from the real world.
With neural textures, we demonstrated a first step towards this avenue, showing applications for photo-realistic novel view point synthesis, scene editing, as well as animation synthesis.
We hope to inspire follow-up work in this direction.

\newpage
\section{Acknowledgments}
We thank Angela Dai for the video voice over and Peter Hedman for the comparison.
We gratefully acknowledge the support by the AI Foundation, Google, a TUM-IAS Rudolf M\"o{\ss}bauer Fellowship, the ERC Starting Grant \textit{Scan2CAD} (804724), and a Google Faculty Award.
This work was supported by the Max Planck Center for Visual Computing and Communications (MPC-VCC).

\newpage
\appendix

\section{Network Architecture}
\label{sec:net_architecutre}
\begin{figure}[h]
	\centering
	\includegraphics[width=\linewidth]{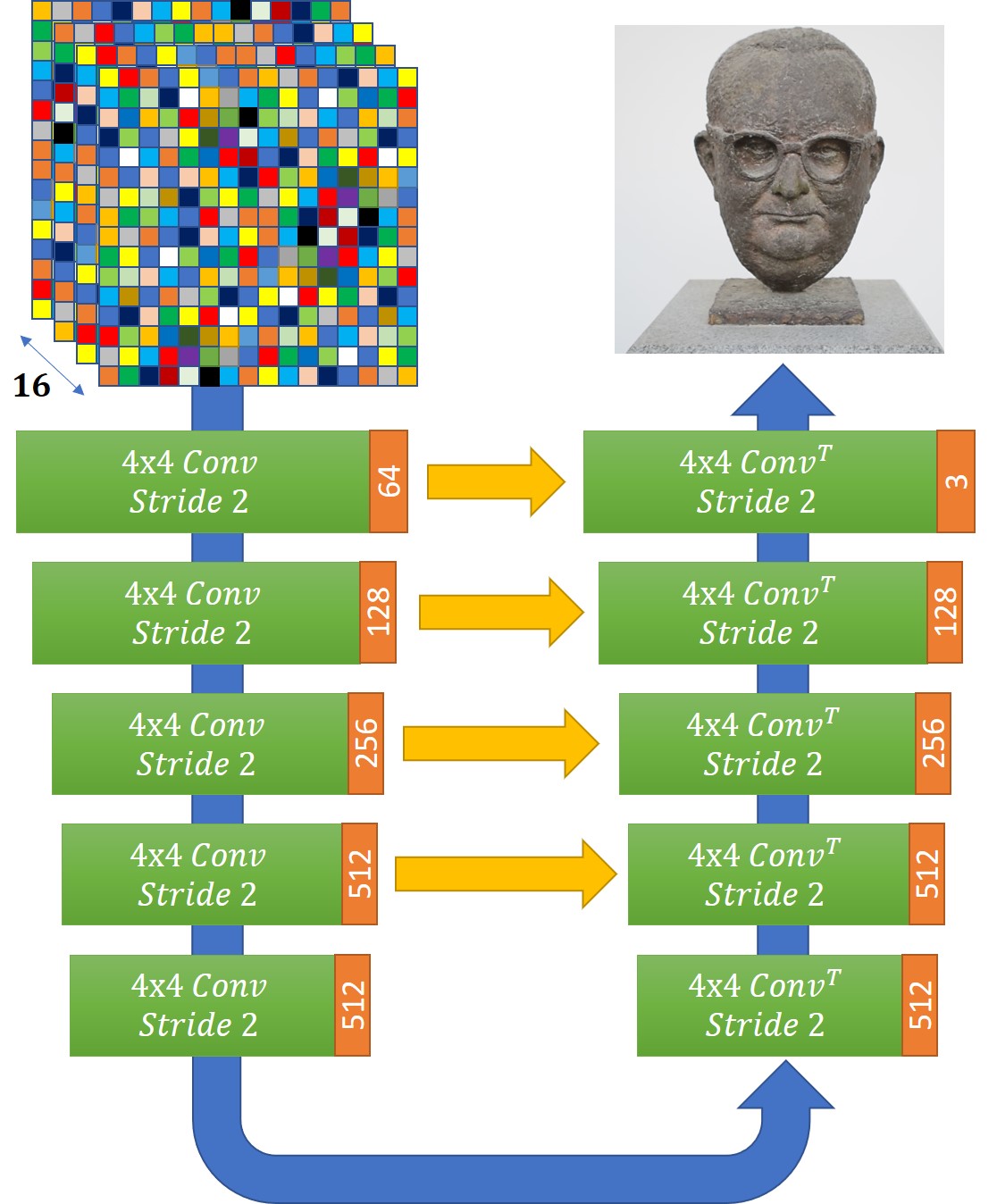}
	\caption
	{
        U-Net architecture of our rendering network. Given the rendered neural textures, we run an encoder-decoder network with skip connections (in yellow), to generate the final output image. Number of output features of the layers are noted in orange.
	}
	\label{fig:net}
\end{figure}

Our rendering network is based on a U-Net~\cite{IsolaZZE2017} with 5-layers, i.e., an encoder-decoder network with skip connections.
For our experiments, we are using the following architecture (see Fig.~\ref{fig:net}).
Based on an image containing $16$ features per pixel (i.e., the rendered neural texture), we apply an encoder that is based on 5 convolutional layers each with instance normalization and a leaky ReLU activation (negative slope of 0.2).
The kernel size of each convolutional layer is $4$ (stride of $2$) with output features $64$ for the first and $128$, $256$, $512$, $512$ respectively for the other layers.
The decoder mirrors the encoder, i.e., the feature channels are the same as in the respective encoder layer, kernel size is $4$ and stride is $2$.
For the final output layer, we are using a $TanH$ activation as in Pix2Pix~\cite{IsolaZZE2017}.

\bibliographystyle{ACM-Reference-Format}
\bibliography{paper_arxiv}


\begin{thebibliography}{69}


\ifx \showCODEN    \undefined \def \showCODEN     #1{\unskip}     \fi
\ifx \showDOI      \undefined \def \showDOI       #1{#1}\fi
\ifx \showISBNx    \undefined \def \showISBNx     #1{\unskip}     \fi
\ifx \showISBNxiii \undefined \def \showISBNxiii  #1{\unskip}     \fi
\ifx \showISSN     \undefined \def \showISSN      #1{\unskip}     \fi
\ifx \showLCCN     \undefined \def \showLCCN      #1{\unskip}     \fi
\ifx \shownote     \undefined \def \shownote      #1{#1}          \fi
\ifx \showarticletitle \undefined \def \showarticletitle #1{#1}   \fi
\ifx \showURL      \undefined \def \showURL       {\relax}        \fi
\providecommand\bibfield[2]{#2}
\providecommand\bibinfo[2]{#2}
\providecommand\natexlab[1]{#1}
\providecommand\showeprint[2][]{arXiv:#2}

\bibitem[\protect\citeauthoryear{Buehler, Bosse, McMillan, Gortler, and
  Cohen}{Buehler et~al\mbox{.}}{2001}]%
        {Buehler:2001}
\bibfield{author}{\bibinfo{person}{Chris Buehler}, \bibinfo{person}{Michael
  Bosse}, \bibinfo{person}{Leonard McMillan}, \bibinfo{person}{Steven Gortler},
  {and} \bibinfo{person}{Michael Cohen}.} \bibinfo{year}{2001}\natexlab{}.
\newblock \showarticletitle{Unstructured Lumigraph Rendering}. In
  \bibinfo{booktitle}{\emph{Proceedings of the 28th Annual Conference on
  Computer Graphics and Interactive Techniques}}
  \emph{(\bibinfo{series}{SIGGRAPH '01})}. \bibinfo{publisher}{ACM},
  \bibinfo{address}{New York, NY, USA}, \bibinfo{pages}{425--432}.
\newblock
\showISBNx{1-58113-374-X}
\urldef\tempurl%
\url{https://doi.org/10.1145/383259.383309}
\showDOI{\tempurl}


\bibitem[\protect\citeauthoryear{Carranza, Theobalt, Magnor, and
  Seidel}{Carranza et~al\mbox{.}}{2003}]%
        {Carranza:2003}
\bibfield{author}{\bibinfo{person}{Joel Carranza}, \bibinfo{person}{Christian
  Theobalt}, \bibinfo{person}{Marcus~A. Magnor}, {and}
  \bibinfo{person}{Hans-Peter Seidel}.} \bibinfo{year}{2003}\natexlab{}.
\newblock \showarticletitle{Free-viewpoint Video of Human Actors}.
\newblock \bibinfo{journal}{\emph{ACM Trans. Graph. (Proc. SIGGRAPH)}}
  \bibinfo{volume}{22}, \bibinfo{number}{3} (\bibinfo{date}{July}
  \bibinfo{year}{2003}), \bibinfo{pages}{569--577}.
\newblock
\showISSN{0730-0301}


\bibitem[\protect\citeauthoryear{Casas, Richardt, Collomosse, Theobalt, and
  Hilton}{Casas et~al\mbox{.}}{2015}]%
        {CasasRCTH15}
\bibfield{author}{\bibinfo{person}{Dan Casas}, \bibinfo{person}{Christian
  Richardt}, \bibinfo{person}{John~P. Collomosse}, \bibinfo{person}{Christian
  Theobalt}, {and} \bibinfo{person}{Adrian Hilton}.}
  \bibinfo{year}{2015}\natexlab{}.
\newblock \showarticletitle{4D Model Flow: Precomputed Appearance Alignment for
  Real-time 4D Video Interpolation}.
\newblock \bibinfo{journal}{\emph{Comput. Graph. Forum}} \bibinfo{volume}{34},
  \bibinfo{number}{7} (\bibinfo{year}{2015}), \bibinfo{pages}{173--182}.
\newblock


\bibitem[\protect\citeauthoryear{Chaurasia, Duchene, Sorkine-Hornung, and
  Drettakis}{Chaurasia et~al\mbox{.}}{2013}]%
        {Chaurasia:2013}
\bibfield{author}{\bibinfo{person}{Gaurav Chaurasia}, \bibinfo{person}{Sylvain
  Duchene}, \bibinfo{person}{Olga Sorkine-Hornung}, {and}
  \bibinfo{person}{George Drettakis}.} \bibinfo{year}{2013}\natexlab{}.
\newblock \showarticletitle{Depth Synthesis and Local Warps for Plausible
  Image-based Navigation}.
\newblock \bibinfo{journal}{\emph{ACM Trans. Graph.}} \bibinfo{volume}{32},
  \bibinfo{number}{3}, Article \bibinfo{articleno}{30} (\bibinfo{date}{July}
  \bibinfo{year}{2013}), \bibinfo{numpages}{12}~pages.
\newblock
\showISSN{0730-0301}


\bibitem[\protect\citeauthoryear{Chen, Wu, Zhang, Li, Lu, Gao, and Yu}{Chen
  et~al\mbox{.}}{2018}]%
        {Chen2018}
\bibfield{author}{\bibinfo{person}{Anpei Chen}, \bibinfo{person}{Minye Wu},
  \bibinfo{person}{Yingliang Zhang}, \bibinfo{person}{Nianyi Li},
  \bibinfo{person}{Jie Lu}, \bibinfo{person}{Shenghua Gao}, {and}
  \bibinfo{person}{Jingyi Yu}.} \bibinfo{year}{2018}\natexlab{}.
\newblock \showarticletitle{Deep Surface Light Fields}.
\newblock \bibinfo{journal}{\emph{Proc. ACM Comput. Graph. Interact. Tech.}}
  \bibinfo{volume}{1}, \bibinfo{number}{1}, Article \bibinfo{articleno}{14}
  (\bibinfo{date}{July} \bibinfo{year}{2018}), \bibinfo{numpages}{17}~pages.
\newblock
\showISSN{2577-6193}
\urldef\tempurl%
\url{https://doi.org/10.1145/3203192}
\showDOI{\tempurl}


\bibitem[\protect\citeauthoryear{Chen, Bautembach, and Izadi}{Chen
  et~al\mbox{.}}{2013}]%
        {chen2013scalable}
\bibfield{author}{\bibinfo{person}{Jiawen Chen}, \bibinfo{person}{Dennis
  Bautembach}, {and} \bibinfo{person}{Shahram Izadi}.}
  \bibinfo{year}{2013}\natexlab{}.
\newblock \showarticletitle{Scalable real-time volumetric surface
  reconstruction}.
\newblock \bibinfo{journal}{\emph{ACM Transactions on Graphics (ToG)}}
  \bibinfo{volume}{32}, \bibinfo{number}{4} (\bibinfo{year}{2013}),
  \bibinfo{pages}{113}.
\newblock


\bibitem[\protect\citeauthoryear{Chen, Bouguet, Chu, and Grzeszczuk}{Chen
  et~al\mbox{.}}{2002}]%
        {Chen2002}
\bibfield{author}{\bibinfo{person}{Wei-Chao Chen}, \bibinfo{person}{Jean-Yves
  Bouguet}, \bibinfo{person}{Michael~H. Chu}, {and} \bibinfo{person}{Radek
  Grzeszczuk}.} \bibinfo{year}{2002}\natexlab{}.
\newblock \showarticletitle{Light Field Mapping: Efficient Representation and
  Hardware Rendering of Surface Light Fields}. In
  \bibinfo{booktitle}{\emph{Proceedings of the 29th Annual Conference on
  Computer Graphics and Interactive Techniques}}
  \emph{(\bibinfo{series}{SIGGRAPH '02})}. \bibinfo{publisher}{ACM},
  \bibinfo{address}{New York, NY, USA}, \bibinfo{pages}{447--456}.
\newblock
\showISBNx{1-58113-521-1}
\urldef\tempurl%
\url{https://doi.org/10.1145/566570.566601}
\showDOI{\tempurl}


\bibitem[\protect\citeauthoryear{Choi, Zhou, and Koltun}{Choi
  et~al\mbox{.}}{2015}]%
        {choi2015robust}
\bibfield{author}{\bibinfo{person}{Sungjoon Choi}, \bibinfo{person}{Qian-Yi
  Zhou}, {and} \bibinfo{person}{Vladlen Koltun}.}
  \bibinfo{year}{2015}\natexlab{}.
\newblock \showarticletitle{Robust reconstruction of indoor scenes}. In
  \bibinfo{booktitle}{\emph{Proceedings of the IEEE Conference on Computer
  Vision and Pattern Recognition}}. \bibinfo{pages}{5556--5565}.
\newblock


\bibitem[\protect\citeauthoryear{Cohen and Welling}{Cohen and Welling}{2014}]%
        {cohen2014transformation}
\bibfield{author}{\bibinfo{person}{Taco~S Cohen} {and} \bibinfo{person}{Max
  Welling}.} \bibinfo{year}{2014}\natexlab{}.
\newblock \showarticletitle{Transformation properties of learned visual
  representations}.
\newblock \bibinfo{journal}{\emph{arXiv preprint arXiv:1412.7659}}
  (\bibinfo{year}{2014}).
\newblock


\bibitem[\protect\citeauthoryear{Cozzolino, Thies, R\"ossler, Riess,
  Nie{\ss}ner, and Verdoliva}{Cozzolino et~al\mbox{.}}{2018}]%
        {cozzolino2018forensictransfer}
\bibfield{author}{\bibinfo{person}{Davide Cozzolino}, \bibinfo{person}{Justus
  Thies}, \bibinfo{person}{Andreas R\"ossler}, \bibinfo{person}{Christian
  Riess}, \bibinfo{person}{Matthias Nie{\ss}ner}, {and} \bibinfo{person}{Luisa
  Verdoliva}.} \bibinfo{year}{2018}\natexlab{}.
\newblock \showarticletitle{ForensicTransfer: Weakly-supervised Domain
  Adaptation for Forgery Detection}.
\newblock \bibinfo{journal}{\emph{arXiv}} (\bibinfo{year}{2018}).
\newblock


\bibitem[\protect\citeauthoryear{Dai, Nie{\ss}ner, Zollh{\"o}fer, Izadi, and
  Theobalt}{Dai et~al\mbox{.}}{2017}]%
        {dai2017bundlefusion}
\bibfield{author}{\bibinfo{person}{Angela Dai}, \bibinfo{person}{Matthias
  Nie{\ss}ner}, \bibinfo{person}{Michael Zollh{\"o}fer},
  \bibinfo{person}{Shahram Izadi}, {and} \bibinfo{person}{Christian Theobalt}.}
  \bibinfo{year}{2017}\natexlab{}.
\newblock \showarticletitle{Bundlefusion: Real-time globally consistent 3d
  reconstruction using on-the-fly surface reintegration}.
\newblock \bibinfo{journal}{\emph{ACM Transactions on Graphics (TOG)}}
  \bibinfo{volume}{36}, \bibinfo{number}{4} (\bibinfo{year}{2017}),
  \bibinfo{pages}{76a}.
\newblock


\bibitem[\protect\citeauthoryear{Debevec, Yu, and Boshokov}{Debevec
  et~al\mbox{.}}{1998}]%
        {Debevec1998}
\bibfield{author}{\bibinfo{person}{Paul Debevec}, \bibinfo{person}{Yizhou Yu},
  {and} \bibinfo{person}{George Boshokov}.} \bibinfo{year}{1998}\natexlab{}.
\newblock \showarticletitle{Efficient View-Dependent {IBR} with Projective
  Texture-Mapping}. \bibinfo{publisher}{EG Rendering Workshop}.
\newblock


\bibitem[\protect\citeauthoryear{Dou, Khamis, Degtyarev, Davidson, Fanello,
  Kowdle, Escolano, Rhemann, Kim, Taylor, et~al\mbox{.}}{Dou
  et~al\mbox{.}}{2016}]%
        {dou2016fusion4d}
\bibfield{author}{\bibinfo{person}{Mingsong Dou}, \bibinfo{person}{Sameh
  Khamis}, \bibinfo{person}{Yury Degtyarev}, \bibinfo{person}{Philip Davidson},
  \bibinfo{person}{Sean~Ryan Fanello}, \bibinfo{person}{Adarsh Kowdle},
  \bibinfo{person}{Sergio~Orts Escolano}, \bibinfo{person}{Christoph Rhemann},
  \bibinfo{person}{David Kim}, \bibinfo{person}{Jonathan Taylor},
  {et~al\mbox{.}}} \bibinfo{year}{2016}\natexlab{}.
\newblock \showarticletitle{Fusion4d: Real-time performance capture of
  challenging scenes}.
\newblock \bibinfo{journal}{\emph{ACM Transactions on Graphics (TOG)}}
  \bibinfo{volume}{35}, \bibinfo{number}{4} (\bibinfo{year}{2016}),
  \bibinfo{pages}{114}.
\newblock


\bibitem[\protect\citeauthoryear{Du, Chuang, Chang, Hoppe, and Varshney}{Du
  et~al\mbox{.}}{2018}]%
        {Montage4D}
\bibfield{author}{\bibinfo{person}{Ruofei Du}, \bibinfo{person}{Ming Chuang},
  \bibinfo{person}{Wayne Chang}, \bibinfo{person}{Hugues Hoppe}, {and}
  \bibinfo{person}{Amitabh Varshney}.} \bibinfo{year}{2018}\natexlab{}.
\newblock \showarticletitle{Montage4D: Interactive Seamless Fusion of Multiview
  Video Textures}. In \bibinfo{booktitle}{\emph{Proceedings of the ACM SIGGRAPH
  Symposium on Interactive 3D Graphics and Games}} \emph{(\bibinfo{series}{I3D
  '18})}. \bibinfo{publisher}{ACM}, \bibinfo{address}{New York, NY, USA},
  Article \bibinfo{articleno}{5}, \bibinfo{numpages}{11}~pages.
\newblock
\showISBNx{978-1-4503-5705-0}
\urldef\tempurl%
\url{https://doi.org/10.1145/3190834.3190843}
\showDOI{\tempurl}


\bibitem[\protect\citeauthoryear{Eisemann, De~Decker, Magnor, Bekaert,
  De~Aguiar, Ahmed, Theobalt, and Sellent}{Eisemann et~al\mbox{.}}{2008}]%
        {EisemannFT}
\bibfield{author}{\bibinfo{person}{M. Eisemann}, \bibinfo{person}{B.
  De~Decker}, \bibinfo{person}{M. Magnor}, \bibinfo{person}{P. Bekaert},
  \bibinfo{person}{E. De~Aguiar}, \bibinfo{person}{N. Ahmed},
  \bibinfo{person}{C. Theobalt}, {and} \bibinfo{person}{A. Sellent}.}
  \bibinfo{year}{2008}\natexlab{}.
\newblock \showarticletitle{{Floating Textures}}.
\newblock \bibinfo{journal}{\emph{Computer Graphics Forum (Proc. EUROGRAHICS}}
  (\bibinfo{year}{2008}).
\newblock
\showISSN{1467-8659}
\urldef\tempurl%
\url{https://doi.org/10.1111/j.1467-8659.2008.01138.x}
\showDOI{\tempurl}


\bibitem[\protect\citeauthoryear{Eslami, Rezende, Besse, Viola, Morcos,
  Garnelo, Ruderman, Rusu, Danihelka, Gregor, et~al\mbox{.}}{Eslami
  et~al\mbox{.}}{2018}]%
        {eslami2018neural}
\bibfield{author}{\bibinfo{person}{SM~Ali Eslami},
  \bibinfo{person}{Danilo~Jimenez Rezende}, \bibinfo{person}{Frederic Besse},
  \bibinfo{person}{Fabio Viola}, \bibinfo{person}{Ari~S Morcos},
  \bibinfo{person}{Marta Garnelo}, \bibinfo{person}{Avraham Ruderman},
  \bibinfo{person}{Andrei~A Rusu}, \bibinfo{person}{Ivo Danihelka},
  \bibinfo{person}{Karol Gregor}, {et~al\mbox{.}}}
  \bibinfo{year}{2018}\natexlab{}.
\newblock \showarticletitle{Neural scene representation and rendering}.
\newblock \bibinfo{journal}{\emph{Science}} \bibinfo{volume}{360},
  \bibinfo{number}{6394} (\bibinfo{year}{2018}), \bibinfo{pages}{1204--1210}.
\newblock


\bibitem[\protect\citeauthoryear{Falorsi, de~Haan, Davidson, De~Cao, Weiler,
  Forr{\'e}, and Cohen}{Falorsi et~al\mbox{.}}{2018}]%
        {falorsi2018explorations}
\bibfield{author}{\bibinfo{person}{Luca Falorsi}, \bibinfo{person}{Pim de
  Haan}, \bibinfo{person}{Tim~R Davidson}, \bibinfo{person}{Nicola De~Cao},
  \bibinfo{person}{Maurice Weiler}, \bibinfo{person}{Patrick Forr{\'e}}, {and}
  \bibinfo{person}{Taco~S Cohen}.} \bibinfo{year}{2018}\natexlab{}.
\newblock \showarticletitle{Explorations in Homeomorphic Variational
  Auto-Encoding}.
\newblock \bibinfo{journal}{\emph{ICML Workshop on Theoretical Foundations and
  Applications of Generative Models}} (\bibinfo{year}{2018}).
\newblock


\bibitem[\protect\citeauthoryear{Flynn, Neulander, Philbin, and Snavely}{Flynn
  et~al\mbox{.}}{2016}]%
        {Flynn2016}
\bibfield{author}{\bibinfo{person}{John Flynn}, \bibinfo{person}{Ivan
  Neulander}, \bibinfo{person}{James Philbin}, {and} \bibinfo{person}{Noah
  Snavely}.} \bibinfo{year}{2016}\natexlab{}.
\newblock \showarticletitle{Deepstereo: Learning to predict new views from the
  world's imagery}. In \bibinfo{booktitle}{\emph{Proc. CVPR}}.
  \bibinfo{pages}{5515--5524}.
\newblock


\bibitem[\protect\citeauthoryear{Goodfellow, Pouget-Abadie, Mirza, Xu,
  Warde-Farley, Ozair, Courville, and Bengio}{Goodfellow et~al\mbox{.}}{2014}]%
        {goodfellow2014generative}
\bibfield{author}{\bibinfo{person}{Ian Goodfellow}, \bibinfo{person}{Jean
  Pouget-Abadie}, \bibinfo{person}{Mehdi Mirza}, \bibinfo{person}{Bing Xu},
  \bibinfo{person}{David Warde-Farley}, \bibinfo{person}{Sherjil Ozair},
  \bibinfo{person}{Aaron Courville}, {and} \bibinfo{person}{Yoshua Bengio}.}
  \bibinfo{year}{2014}\natexlab{}.
\newblock \showarticletitle{Generative adversarial nets}. In
  \bibinfo{booktitle}{\emph{Proc. NIPS}}. \bibinfo{pages}{2672--2680}.
\newblock


\bibitem[\protect\citeauthoryear{Gortler, Grzeszczuk, Szeliski, and
  Cohen}{Gortler et~al\mbox{.}}{1996}]%
        {Gortler1996}
\bibfield{author}{\bibinfo{person}{Steven~J. Gortler}, \bibinfo{person}{Radek
  Grzeszczuk}, \bibinfo{person}{Richard Szeliski}, {and}
  \bibinfo{person}{Michael~F. Cohen}.} \bibinfo{year}{1996}\natexlab{}.
\newblock \showarticletitle{The Lumigraph}. In
  \bibinfo{booktitle}{\emph{Proceedings of the 23rd Annual Conference on
  Computer Graphics and Interactive Techniques}}
  \emph{(\bibinfo{series}{SIGGRAPH '96})}. \bibinfo{publisher}{ACM},
  \bibinfo{address}{New York, NY, USA}, \bibinfo{pages}{43--54}.
\newblock
\showISBNx{0-89791-746-4}
\urldef\tempurl%
\url{https://doi.org/10.1145/237170.237200}
\showDOI{\tempurl}


\bibitem[\protect\citeauthoryear{Hedman, Philip, Price, Frahm, Drettakis, and
  Brostow}{Hedman et~al\mbox{.}}{2018}]%
        {hedman2018deep}
\bibfield{author}{\bibinfo{person}{Peter Hedman}, \bibinfo{person}{Julien
  Philip}, \bibinfo{person}{True Price}, \bibinfo{person}{Jan-Michael Frahm},
  \bibinfo{person}{George Drettakis}, {and} \bibinfo{person}{Gabriel Brostow}.}
  \bibinfo{year}{2018}\natexlab{}.
\newblock \showarticletitle{Deep blending for free-viewpoint image-based
  rendering}. In \bibinfo{booktitle}{\emph{SIGGRAPH Asia 2018 Technical
  Papers}}. ACM, \bibinfo{pages}{257}.
\newblock


\bibitem[\protect\citeauthoryear{Hedman, Ritschel, Drettakis, and
  Brostow}{Hedman et~al\mbox{.}}{2016}]%
        {hedman2016scalable}
\bibfield{author}{\bibinfo{person}{Peter Hedman}, \bibinfo{person}{Tobias
  Ritschel}, \bibinfo{person}{George Drettakis}, {and} \bibinfo{person}{Gabriel
  Brostow}.} \bibinfo{year}{2016}\natexlab{}.
\newblock \showarticletitle{Scalable inside-out image-based rendering}.
\newblock \bibinfo{journal}{\emph{ACM Transactions on Graphics (TOG)}}
  \bibinfo{volume}{35}, \bibinfo{number}{6} (\bibinfo{year}{2016}),
  \bibinfo{pages}{231}.
\newblock


\bibitem[\protect\citeauthoryear{Heigl, Koch, Pollefeys, Denzler, and
  Gool}{Heigl et~al\mbox{.}}{1999}]%
        {HeiglDAGM99}
\bibfield{author}{\bibinfo{person}{Benno Heigl}, \bibinfo{person}{Reinhard
  Koch}, \bibinfo{person}{Marc Pollefeys}, \bibinfo{person}{Joachim Denzler},
  {and} \bibinfo{person}{Luc J.~Van Gool}.} \bibinfo{year}{1999}\natexlab{}.
\newblock \showarticletitle{Plenoptic Modeling and Rendering from Image
  Sequences Taken by Hand-Held Camera}. In \bibinfo{booktitle}{\emph{Proc.
  DAGM}}. \bibinfo{pages}{94--101}.
\newblock


\bibitem[\protect\citeauthoryear{Hinton and Salakhutdinov}{Hinton and
  Salakhutdinov}{2006}]%
        {HintoS2006}
\bibfield{author}{\bibinfo{person}{Geoffrey~E. Hinton} {and}
  \bibinfo{person}{Ruslan Salakhutdinov}.} \bibinfo{year}{2006}\natexlab{}.
\newblock \showarticletitle{Reducing the Dimensionality of Data with Neural
  Networks}.
\newblock \bibinfo{journal}{\emph{Science}} \bibinfo{volume}{313},
  \bibinfo{number}{5786} (\bibinfo{date}{July} \bibinfo{year}{2006}),
  \bibinfo{pages}{504--507}.
\newblock
\showISSN{0036-8075}
\urldef\tempurl%
\url{https://doi.org/10.1126/science.1127647}
\showDOI{\tempurl}


\bibitem[\protect\citeauthoryear{Huang, Dai, Guibas, and Nie{\ss}ner}{Huang
  et~al\mbox{.}}{2017}]%
        {huang2017dlight}
\bibfield{author}{\bibinfo{person}{Jingwei Huang}, \bibinfo{person}{Angela
  Dai}, \bibinfo{person}{Leonidas Guibas}, {and} \bibinfo{person}{Matthias
  Nie{\ss}ner}.} \bibinfo{year}{2017}\natexlab{}.
\newblock \showarticletitle{3DLite: Towards Commodity 3D Scanning for Content
  Creation}.
\newblock \bibinfo{journal}{\emph{ACM Transactions on Graphics 2017 (TOG)}}
  (\bibinfo{year}{2017}).
\newblock


\bibitem[\protect\citeauthoryear{Innmann, Zollh{\"o}fer, Nie{\ss}ner, Theobalt,
  and Stamminger}{Innmann et~al\mbox{.}}{2016}]%
        {innmann2016volumedeform}
\bibfield{author}{\bibinfo{person}{Matthias Innmann}, \bibinfo{person}{Michael
  Zollh{\"o}fer}, \bibinfo{person}{Matthias Nie{\ss}ner},
  \bibinfo{person}{Christian Theobalt}, {and} \bibinfo{person}{Marc
  Stamminger}.} \bibinfo{year}{2016}\natexlab{}.
\newblock \showarticletitle{VolumeDeform: Real-time volumetric non-rigid
  reconstruction}. In \bibinfo{booktitle}{\emph{European Conference on Computer
  Vision}}. Springer, \bibinfo{pages}{362--379}.
\newblock


\bibitem[\protect\citeauthoryear{Isola, Zhu, Zhou, and Efros}{Isola
  et~al\mbox{.}}{2016}]%
        {pix2pix}
\bibfield{author}{\bibinfo{person}{Phillip Isola}, \bibinfo{person}{Jun-Yan
  Zhu}, \bibinfo{person}{Tinghui Zhou}, {and} \bibinfo{person}{Alexei~A
  Efros}.} \bibinfo{year}{2016}\natexlab{}.
\newblock \showarticletitle{Image-to-Image Translation with Conditional
  Adversarial Networks}.
\newblock \bibinfo{journal}{\emph{arxiv}} (\bibinfo{year}{2016}).
\newblock


\bibitem[\protect\citeauthoryear{Isola, Zhu, Zhou, and Efros}{Isola
  et~al\mbox{.}}{2017}]%
        {IsolaZZE2017}
\bibfield{author}{\bibinfo{person}{Phillip Isola}, \bibinfo{person}{Jun-Yan
  Zhu}, \bibinfo{person}{Tinghui Zhou}, {and} \bibinfo{person}{Alexei~A.
  Efros}.} \bibinfo{year}{2017}\natexlab{}.
\newblock \showarticletitle{Image-to-Image Translation with Conditional
  Adversarial Networks}. In \bibinfo{booktitle}{\emph{Proc. CVPR}}.
  \bibinfo{pages}{5967--5976}.
\newblock
\urldef\tempurl%
\url{https://doi.org/10.1109/CVPR.2017.632}
\showDOI{\tempurl}


\bibitem[\protect\citeauthoryear{Izadi, Kim, Hilliges, Molyneaux, Newcombe,
  Kohli, Shotton, Hodges, Freeman, Davison, et~al\mbox{.}}{Izadi
  et~al\mbox{.}}{2011}]%
        {izadi2011kinectfusion}
\bibfield{author}{\bibinfo{person}{Shahram Izadi}, \bibinfo{person}{David Kim},
  \bibinfo{person}{Otmar Hilliges}, \bibinfo{person}{David Molyneaux},
  \bibinfo{person}{Richard Newcombe}, \bibinfo{person}{Pushmeet Kohli},
  \bibinfo{person}{Jamie Shotton}, \bibinfo{person}{Steve Hodges},
  \bibinfo{person}{Dustin Freeman}, \bibinfo{person}{Andrew Davison},
  {et~al\mbox{.}}} \bibinfo{year}{2011}\natexlab{}.
\newblock \showarticletitle{KinectFusion: real-time 3D reconstruction and
  interaction using a moving depth camera}. In
  \bibinfo{booktitle}{\emph{Proceedings of the 24th annual ACM symposium on
  User interface software and technology}}. ACM, \bibinfo{pages}{559--568}.
\newblock


\bibitem[\protect\citeauthoryear{Kalantari, Wang, and Ramamoorthi}{Kalantari
  et~al\mbox{.}}{2016}]%
        {LearningViewSynthesis}
\bibfield{author}{\bibinfo{person}{Nima~Khademi Kalantari},
  \bibinfo{person}{Ting-Chun Wang}, {and} \bibinfo{person}{Ravi Ramamoorthi}.}
  \bibinfo{year}{2016}\natexlab{}.
\newblock \showarticletitle{Learning-Based View Synthesis for Light Field
  Cameras}.
\newblock \bibinfo{journal}{\emph{ACM Transactions on Graphics (Proceedings of
  SIGGRAPH Asia 2016)}} \bibinfo{volume}{35}, \bibinfo{number}{6}
  (\bibinfo{year}{2016}).
\newblock


\bibitem[\protect\citeauthoryear{Karras, Aila, Laine, and Lehtinen}{Karras
  et~al\mbox{.}}{2018}]%
        {KarraALL2018}
\bibfield{author}{\bibinfo{person}{Tero Karras}, \bibinfo{person}{Timo Aila},
  \bibinfo{person}{Samuli Laine}, {and} \bibinfo{person}{Jaakko Lehtinen}.}
  \bibinfo{year}{2018}\natexlab{}.
\newblock \showarticletitle{Progressive Growing of {GANs} for Improved Quality,
  Stability, and Variation}. In \bibinfo{booktitle}{\emph{Proc. ICLR}}.
\newblock


\bibitem[\protect\citeauthoryear{Kim, Garrido, Tewari, Xu, Thies, Nie{\ss}ner,
  P{\'e}rez, Richardt, Zollh{\"o}fer, and Theobalt}{Kim et~al\mbox{.}}{2018}]%
        {kim2018DeepVideo}
\bibfield{author}{\bibinfo{person}{H. Kim}, \bibinfo{person}{P. Garrido},
  \bibinfo{person}{A. Tewari}, \bibinfo{person}{W. Xu}, \bibinfo{person}{J.
  Thies}, \bibinfo{person}{N. Nie{\ss}ner}, \bibinfo{person}{P. P{\'e}rez},
  \bibinfo{person}{C. Richardt}, \bibinfo{person}{M. Zollh{\"o}fer}, {and}
  \bibinfo{person}{C. Theobalt}.} \bibinfo{year}{2018}\natexlab{}.
\newblock \showarticletitle{{Deep Video Portraits}}.
\newblock \bibinfo{journal}{\emph{ACM Transactions on Graphics 2018 (TOG)}}
  (\bibinfo{year}{2018}).
\newblock


\bibitem[\protect\citeauthoryear{Kingma and Ba}{Kingma and Ba}{2014}]%
        {adam}
\bibfield{author}{\bibinfo{person}{Diederik~P. Kingma} {and}
  \bibinfo{person}{Jimmy Ba}.} \bibinfo{year}{2014}\natexlab{}.
\newblock \showarticletitle{Adam: {A} Method for Stochastic Optimization}.
\newblock \bibinfo{journal}{\emph{CoRR}}  \bibinfo{volume}{abs/1412.6980}
  (\bibinfo{year}{2014}).
\newblock
\showeprint[arxiv]{1412.6980}
\urldef\tempurl%
\url{http://arxiv.org/abs/1412.6980}
\showURL{%
\tempurl}


\bibitem[\protect\citeauthoryear{Kingma and Welling}{Kingma and
  Welling}{2013}]%
        {jKingma2014}
\bibfield{author}{\bibinfo{person}{Diederik~P. Kingma} {and}
  \bibinfo{person}{Max Welling}.} \bibinfo{year}{2013}\natexlab{}.
\newblock \showarticletitle{Auto-Encoding Variational Bayes.}
\newblock \bibinfo{journal}{\emph{CoRR}}  \bibinfo{volume}{abs/1312.6114}
  (\bibinfo{year}{2013}).
\newblock


\bibitem[\protect\citeauthoryear{Kulkarni, Whitney, Kohli, and
  Tenenbaum}{Kulkarni et~al\mbox{.}}{2015}]%
        {Kulkarni2015}
\bibfield{author}{\bibinfo{person}{Tejas~D Kulkarni},
  \bibinfo{person}{William~F. Whitney}, \bibinfo{person}{Pushmeet Kohli}, {and}
  \bibinfo{person}{Josh Tenenbaum}.} \bibinfo{year}{2015}\natexlab{}.
\newblock \showarticletitle{Deep Convolutional Inverse Graphics Network}.
\newblock In \bibinfo{booktitle}{\emph{Proc. NIPS}}.
  \bibinfo{pages}{2539--2547}.
\newblock


\bibitem[\protect\citeauthoryear{Levoy and Hanrahan}{Levoy and
  Hanrahan}{1996}]%
        {Levoy1996}
\bibfield{author}{\bibinfo{person}{Marc Levoy} {and} \bibinfo{person}{Pat
  Hanrahan}.} \bibinfo{year}{1996}\natexlab{}.
\newblock \showarticletitle{Light Field Rendering}. In
  \bibinfo{booktitle}{\emph{Proceedings of the 23rd Annual Conference on
  Computer Graphics and Interactive Techniques}}
  \emph{(\bibinfo{series}{SIGGRAPH '96})}. \bibinfo{publisher}{ACM},
  \bibinfo{address}{New York, NY, USA}, \bibinfo{pages}{31--42}.
\newblock
\showISBNx{0-89791-746-4}
\urldef\tempurl%
\url{https://doi.org/10.1145/237170.237199}
\showDOI{\tempurl}


\bibitem[\protect\citeauthoryear{Li, Gharbi, Adams, Durand, and
  Ragan-Kelley}{Li et~al\mbox{.}}{2018}]%
        {li2018differentiable}
\bibfield{author}{\bibinfo{person}{Tzu-Mao Li}, \bibinfo{person}{Micha{\"e}l
  Gharbi}, \bibinfo{person}{Andrew Adams}, \bibinfo{person}{Fr{\'e}do Durand},
  {and} \bibinfo{person}{Jonathan Ragan-Kelley}.}
  \bibinfo{year}{2018}\natexlab{}.
\newblock \showarticletitle{Differentiable programming for image processing and
  deep learning in halide}.
\newblock \bibinfo{journal}{\emph{ACM Transactions on Graphics (TOG)}}
  \bibinfo{volume}{37}, \bibinfo{number}{4} (\bibinfo{year}{2018}),
  \bibinfo{pages}{139}.
\newblock


\bibitem[\protect\citeauthoryear{Magnor and Girod}{Magnor and Girod}{1999}]%
        {Magnor99adaptiveblock-based}
\bibfield{author}{\bibinfo{person}{Marcus Magnor} {and} \bibinfo{person}{Bernd
  Girod}.} \bibinfo{year}{1999}\natexlab{}.
\newblock \showarticletitle{Adaptive Block-based Light Field Coding}. In
  \bibinfo{booktitle}{\emph{Proc. 3rd International Workshop on Synthetic and
  Natural Hybrid Coding and Three-Dimensional Imaging IWSNHC3DI'99, Santorini,
  Greece}}. \bibinfo{pages}{140--143}.
\newblock


\bibitem[\protect\citeauthoryear{Miandji, Kronander, and Unger}{Miandji
  et~al\mbox{.}}{2013}]%
        {Miandji2013}
\bibfield{author}{\bibinfo{person}{Ehsan Miandji}, \bibinfo{person}{Joel
  Kronander}, {and} \bibinfo{person}{Jonas Unger}.}
  \bibinfo{year}{2013}\natexlab{}.
\newblock \showarticletitle{Learning Based Compression for Real-time Rendering
  of Surface Light Fields}. In \bibinfo{booktitle}{\emph{ACM SIGGRAPH 2013
  Posters}} \emph{(\bibinfo{series}{SIGGRAPH '13})}. \bibinfo{publisher}{ACM},
  \bibinfo{address}{New York, NY, USA}, Article \bibinfo{articleno}{44},
  \bibinfo{numpages}{1}~pages.
\newblock
\showISBNx{978-1-4503-2342-0}
\urldef\tempurl%
\url{https://doi.org/10.1145/2503385.2503434}
\showDOI{\tempurl}


\bibitem[\protect\citeauthoryear{Mirza and Osindero}{Mirza and
  Osindero}{2014}]%
        {MirzaO2014}
\bibfield{author}{\bibinfo{person}{Mehdi Mirza} {and} \bibinfo{person}{Simon
  Osindero}.} \bibinfo{year}{2014}\natexlab{}.
\newblock \bibinfo{title}{Conditional Generative Adversarial Nets}.
  (\bibinfo{year}{2014}).
\newblock
\urldef\tempurl%
\url{https://arxiv.org/abs/1411.1784}
\showURL{%
\tempurl}
\newblock
\shownote{arXiv:1411.1784.}


\bibitem[\protect\citeauthoryear{Nagano, Seo, Xing, Wei, Li, Saito, Agarwal,
  Fursund, and Li}{Nagano et~al\mbox{.}}{2018}]%
        {pagan}
\bibfield{author}{\bibinfo{person}{Koki Nagano}, \bibinfo{person}{Jaewoo Seo},
  \bibinfo{person}{Jun Xing}, \bibinfo{person}{Lingyu Wei},
  \bibinfo{person}{Zimo Li}, \bibinfo{person}{Shunsuke Saito},
  \bibinfo{person}{Aviral Agarwal}, \bibinfo{person}{Jens Fursund}, {and}
  \bibinfo{person}{Hao Li}.} \bibinfo{year}{2018}\natexlab{}.
\newblock \showarticletitle{paGAN: real-time avatars using dynamic textures}.
  \bibinfo{pages}{1--12}.
\newblock
\urldef\tempurl%
\url{https://doi.org/10.1145/3272127.3275075}
\showDOI{\tempurl}


\bibitem[\protect\citeauthoryear{Newcombe, Fox, and Seitz}{Newcombe
  et~al\mbox{.}}{2015}]%
        {newcombe2015dynamicfusion}
\bibfield{author}{\bibinfo{person}{Richard~A Newcombe}, \bibinfo{person}{Dieter
  Fox}, {and} \bibinfo{person}{Steven~M Seitz}.}
  \bibinfo{year}{2015}\natexlab{}.
\newblock \showarticletitle{Dynamicfusion: Reconstruction and tracking of
  non-rigid scenes in real-time}. In \bibinfo{booktitle}{\emph{Proceedings of
  the IEEE conference on computer vision and pattern recognition}}.
  \bibinfo{pages}{343--352}.
\newblock


\bibitem[\protect\citeauthoryear{Newcombe, Izadi, Hilliges, Molyneaux, Kim,
  Davison, Kohi, Shotton, Hodges, and Fitzgibbon}{Newcombe
  et~al\mbox{.}}{2011}]%
        {newcombe2011kinectfusion}
\bibfield{author}{\bibinfo{person}{Richard~A Newcombe},
  \bibinfo{person}{Shahram Izadi}, \bibinfo{person}{Otmar Hilliges},
  \bibinfo{person}{David Molyneaux}, \bibinfo{person}{David Kim},
  \bibinfo{person}{Andrew~J Davison}, \bibinfo{person}{Pushmeet Kohi},
  \bibinfo{person}{Jamie Shotton}, \bibinfo{person}{Steve Hodges}, {and}
  \bibinfo{person}{Andrew Fitzgibbon}.} \bibinfo{year}{2011}\natexlab{}.
\newblock \showarticletitle{KinectFusion: Real-time dense surface mapping and
  tracking}. In \bibinfo{booktitle}{\emph{Mixed and augmented reality (ISMAR),
  2011 10th IEEE international symposium on}}. IEEE, \bibinfo{pages}{127--136}.
\newblock


\bibitem[\protect\citeauthoryear{Nie{\ss}ner, Zollh{\"o}fer, Izadi, and
  Stamminger}{Nie{\ss}ner et~al\mbox{.}}{2013}]%
        {niessner2013real}
\bibfield{author}{\bibinfo{person}{Matthias Nie{\ss}ner},
  \bibinfo{person}{Michael Zollh{\"o}fer}, \bibinfo{person}{Shahram Izadi},
  {and} \bibinfo{person}{Marc Stamminger}.} \bibinfo{year}{2013}\natexlab{}.
\newblock \showarticletitle{Real-time 3D reconstruction at scale using voxel
  hashing}.
\newblock \bibinfo{journal}{\emph{ACM Transactions on Graphics (ToG)}}
  \bibinfo{volume}{32}, \bibinfo{number}{6} (\bibinfo{year}{2013}),
  \bibinfo{pages}{169}.
\newblock


\bibitem[\protect\citeauthoryear{Oord, Kalchbrenner, Vinyals, Espeholt, Graves,
  and Kavukcuoglu}{Oord et~al\mbox{.}}{2016}]%
        {Oord:2016}
\bibfield{author}{\bibinfo{person}{A\"{a}ron van~den Oord},
  \bibinfo{person}{Nal Kalchbrenner}, \bibinfo{person}{Oriol Vinyals},
  \bibinfo{person}{Lasse Espeholt}, \bibinfo{person}{Alex Graves}, {and}
  \bibinfo{person}{Koray Kavukcuoglu}.} \bibinfo{year}{2016}\natexlab{}.
\newblock \showarticletitle{Conditional Image Generation with PixelCNN
  Decoders}. In \bibinfo{booktitle}{\emph{Proc. NIPS}}.
  \bibinfo{pages}{4797--4805}.
\newblock


\bibitem[\protect\citeauthoryear{Park, Yang, Yumer, Ceylan, and Berg}{Park
  et~al\mbox{.}}{2017}]%
        {Park17}
\bibfield{author}{\bibinfo{person}{Eunbyung Park}, \bibinfo{person}{Jimei
  Yang}, \bibinfo{person}{Ersin Yumer}, \bibinfo{person}{Duygu Ceylan}, {and}
  \bibinfo{person}{Alexander~C. Berg}.} \bibinfo{year}{2017}\natexlab{}.
\newblock \showarticletitle{Transformation-Grounded Image Generation Network
  for Novel 3D View Synthesis}.
\newblock \bibinfo{journal}{\emph{CoRR}}  \bibinfo{volume}{abs/1703.02921}
  (\bibinfo{year}{2017}).
\newblock


\bibitem[\protect\citeauthoryear{Paszke, Gross, Chintala, Chanan, Yang, DeVito,
  Lin, Desmaison, Antiga, and Lerer}{Paszke et~al\mbox{.}}{2017}]%
        {paszke2017automatic}
\bibfield{author}{\bibinfo{person}{Adam Paszke}, \bibinfo{person}{Sam Gross},
  \bibinfo{person}{Soumith Chintala}, \bibinfo{person}{Gregory Chanan},
  \bibinfo{person}{Edward Yang}, \bibinfo{person}{Zachary DeVito},
  \bibinfo{person}{Zeming Lin}, \bibinfo{person}{Alban Desmaison},
  \bibinfo{person}{Luca Antiga}, {and} \bibinfo{person}{Adam Lerer}.}
  \bibinfo{year}{2017}\natexlab{}.
\newblock \showarticletitle{Automatic differentiation in PyTorch}.
\newblock  (\bibinfo{year}{2017}).
\newblock


\bibitem[\protect\citeauthoryear{Penner and Zhang}{Penner and Zhang}{2017}]%
        {Penner:2017}
\bibfield{author}{\bibinfo{person}{Eric Penner} {and} \bibinfo{person}{Li
  Zhang}.} \bibinfo{year}{2017}\natexlab{}.
\newblock \showarticletitle{Soft 3D Reconstruction for View Synthesis}.
\newblock \bibinfo{journal}{\emph{ACM Trans. Graph.}} \bibinfo{volume}{36},
  \bibinfo{number}{6}, Article \bibinfo{articleno}{235} (\bibinfo{date}{Nov.}
  \bibinfo{year}{2017}), \bibinfo{numpages}{11}~pages.
\newblock
\showISSN{0730-0301}


\bibitem[\protect\citeauthoryear{Radford, Metz, and Chintala}{Radford
  et~al\mbox{.}}{2016}]%
        {RadfoMC2016}
\bibfield{author}{\bibinfo{person}{Alec Radford}, \bibinfo{person}{Luke Metz},
  {and} \bibinfo{person}{Soumith Chintala}.} \bibinfo{year}{2016}\natexlab{}.
\newblock \showarticletitle{Unsupervised Representation Learning with Deep
  Convolutional Generative Adversarial Networks}. In
  \bibinfo{booktitle}{\emph{Proc. ICLR}}.
\newblock


\bibitem[\protect\citeauthoryear{Ronneberger, Fischer, and Brox}{Ronneberger
  et~al\mbox{.}}{2015}]%
        {RonneFB2015}
\bibfield{author}{\bibinfo{person}{Olaf Ronneberger}, \bibinfo{person}{Philipp
  Fischer}, {and} \bibinfo{person}{Thomas Brox}.}
  \bibinfo{year}{2015}\natexlab{}.
\newblock \showarticletitle{{U-Net}: Convolutional Networks for Biomedical
  Image Segmentation}. In \bibinfo{booktitle}{\emph{Proc. MICCAI}}.
  \bibinfo{pages}{234--241}.
\newblock
\showISBNx{978-3-319-24574-4}
\urldef\tempurl%
\url{https://doi.org/10.1007/978-3-319-24574-4_28}
\showDOI{\tempurl}


\bibitem[\protect\citeauthoryear{R\"ossler, Cozzolino, Verdoliva, Riess, Thies,
  and Nie{\ss}ner}{R\"ossler et~al\mbox{.}}{2018}]%
        {roessler2018faceforensics}
\bibfield{author}{\bibinfo{person}{Andreas R\"ossler}, \bibinfo{person}{Davide
  Cozzolino}, \bibinfo{person}{Luisa Verdoliva}, \bibinfo{person}{Christian
  Riess}, \bibinfo{person}{Justus Thies}, {and} \bibinfo{person}{Matthias
  Nie{\ss}ner}.} \bibinfo{year}{2018}\natexlab{}.
\newblock \showarticletitle{FaceForensics: A Large-scale Video Dataset for
  Forgery Detection in Human Faces}.
\newblock \bibinfo{journal}{\emph{arXiv}} (\bibinfo{year}{2018}).
\newblock


\bibitem[\protect\citeauthoryear{R\"ossler, Cozzolino, Verdoliva, Riess, Thies,
  and Nie{\ss}ner}{R\"ossler et~al\mbox{.}}{2019}]%
        {roessler2019faceforensics++}
\bibfield{author}{\bibinfo{person}{Andreas R\"ossler}, \bibinfo{person}{Davide
  Cozzolino}, \bibinfo{person}{Luisa Verdoliva}, \bibinfo{person}{Christian
  Riess}, \bibinfo{person}{Justus Thies}, {and} \bibinfo{person}{Matthias
  Nie{\ss}ner}.} \bibinfo{year}{2019}\natexlab{}.
\newblock \showarticletitle{FaceForensics++: Learning to Detect Manipulated
  Facial Images}.
\newblock \bibinfo{journal}{\emph{arXiv}} (\bibinfo{year}{2019}).
\newblock


\bibitem[\protect\citeauthoryear{Sch\"{o}nberger and Frahm}{Sch\"{o}nberger and
  Frahm}{2016}]%
        {colmap}
\bibfield{author}{\bibinfo{person}{Johannes~Lutz Sch\"{o}nberger} {and}
  \bibinfo{person}{Jan-Michael Frahm}.} \bibinfo{year}{2016}\natexlab{}.
\newblock \showarticletitle{{Structure-from-Motion Revisited}}. In
  \bibinfo{booktitle}{\emph{Conference on Computer Vision and Pattern
  Recognition (CVPR)}}.
\newblock


\bibitem[\protect\citeauthoryear{Sch\"{o}nberger, Zheng, Pollefeys, and
  Frahm}{Sch\"{o}nberger et~al\mbox{.}}{2016}]%
        {colmapB}
\bibfield{author}{\bibinfo{person}{Johannes~Lutz Sch\"{o}nberger},
  \bibinfo{person}{Enliang Zheng}, \bibinfo{person}{Marc Pollefeys}, {and}
  \bibinfo{person}{Jan-Michael Frahm}.} \bibinfo{year}{2016}\natexlab{}.
\newblock \showarticletitle{{Pixelwise View Selection for Unstructured
  Multi-View Stereo}}. In \bibinfo{booktitle}{\emph{European Conference on
  Computer Vision (ECCV)}}.
\newblock


\bibitem[\protect\citeauthoryear{Sitzmann, Thies, Heide, Nießner, Wetzstein,
  and Zollhöfer}{Sitzmann et~al\mbox{.}}{2019}]%
        {Sitzmann:2018:DeepVoxels}
\bibfield{author}{\bibinfo{person}{V. Sitzmann}, \bibinfo{person}{J. Thies},
  \bibinfo{person}{F. Heide}, \bibinfo{person}{M. Nießner},
  \bibinfo{person}{G. Wetzstein}, {and} \bibinfo{person}{M. Zollhöfer}.}
  \bibinfo{year}{2019}\natexlab{}.
\newblock \showarticletitle{DeepVoxels: Learning Persistent 3D Feature
  Embeddings}. In \bibinfo{booktitle}{\emph{Proc. Computer Vision and Pattern
  Recognition (CVPR), IEEE}}.
\newblock


\bibitem[\protect\citeauthoryear{Thies, Zollh{\"o}fer, Stamminger, Theobalt,
  and Nie{\ss}ner}{Thies et~al\mbox{.}}{2016}]%
        {thies2016face}
\bibfield{author}{\bibinfo{person}{Justus Thies}, \bibinfo{person}{M.
  Zollh{\"o}fer}, \bibinfo{person}{M. Stamminger}, \bibinfo{person}{C.
  Theobalt}, {and} \bibinfo{person}{M. Nie{\ss}ner}.}
  \bibinfo{year}{2016}\natexlab{}.
\newblock \showarticletitle{{Face2Face: Real-time Face Capture and Reenactment
  of RGB Videos}}. In \bibinfo{booktitle}{\emph{Proc. CVPR}}.
\newblock


\bibitem[\protect\citeauthoryear{Thies, Zollh{\"o}fer, Theobalt, Stamminger,
  and Nie{\ss}ner}{Thies et~al\mbox{.}}{2018}]%
        {thies2018ignor}
\bibfield{author}{\bibinfo{person}{J. Thies}, \bibinfo{person}{M.
  Zollh{\"o}fer}, \bibinfo{person}{C. Theobalt}, \bibinfo{person}{M.
  Stamminger}, {and} \bibinfo{person}{M. Nie{\ss}ner}.}
  \bibinfo{year}{2018}\natexlab{}.
\newblock \showarticletitle{IGNOR: Image-guided Neural Object Rendering}.
\newblock \bibinfo{journal}{\emph{arXiv 2018}} (\bibinfo{year}{2018}).
\newblock


\bibitem[\protect\citeauthoryear{Tulsiani, Tucker, and Snavely}{Tulsiani
  et~al\mbox{.}}{2018}]%
        {lsiTulsiani18}
\bibfield{author}{\bibinfo{person}{Shubham Tulsiani}, \bibinfo{person}{Richard
  Tucker}, {and} \bibinfo{person}{Noah Snavely}.}
  \bibinfo{year}{2018}\natexlab{}.
\newblock \showarticletitle{Layer-structured 3D Scene Inference via View
  Synthesis}. In \bibinfo{booktitle}{\emph{Proc. ECCV}}.
\newblock


\bibitem[\protect\citeauthoryear{Wang, Liu, Zhu, Liu, Tao, Kautz, and
  Catanzaro}{Wang et~al\mbox{.}}{2018a}]%
        {wang2018vid2vid}
\bibfield{author}{\bibinfo{person}{Ting-Chun Wang}, \bibinfo{person}{Ming-Yu
  Liu}, \bibinfo{person}{Jun-Yan Zhu}, \bibinfo{person}{Guilin Liu},
  \bibinfo{person}{Andrew Tao}, \bibinfo{person}{Jan Kautz}, {and}
  \bibinfo{person}{Bryan Catanzaro}.} \bibinfo{year}{2018}\natexlab{a}.
\newblock \showarticletitle{Video-to-Video Synthesis}. In
  \bibinfo{booktitle}{\emph{Proc. NeurIPS}}.
\newblock


\bibitem[\protect\citeauthoryear{Wang, Liu, Zhu, Tao, Kautz, and
  Catanzaro}{Wang et~al\mbox{.}}{2018b}]%
        {WangLZTKC2018}
\bibfield{author}{\bibinfo{person}{Ting-Chun Wang}, \bibinfo{person}{Ming-Yu
  Liu}, \bibinfo{person}{Jun-Yan Zhu}, \bibinfo{person}{Andrew Tao},
  \bibinfo{person}{Jan Kautz}, {and} \bibinfo{person}{Bryan Catanzaro}.}
  \bibinfo{year}{2018}\natexlab{b}.
\newblock \showarticletitle{High-Resolution Image Synthesis and Semantic
  Manipulation with Conditional {GANs}}. In \bibinfo{booktitle}{\emph{Proc.
  CVPR}}.
\newblock


\bibitem[\protect\citeauthoryear{Whelan, Salas-Moreno, Glocker, Davison, and
  Leutenegger}{Whelan et~al\mbox{.}}{2016}]%
        {whelan2016elasticfusion}
\bibfield{author}{\bibinfo{person}{Thomas Whelan}, \bibinfo{person}{Renato~F
  Salas-Moreno}, \bibinfo{person}{Ben Glocker}, \bibinfo{person}{Andrew~J
  Davison}, {and} \bibinfo{person}{Stefan Leutenegger}.}
  \bibinfo{year}{2016}\natexlab{}.
\newblock \showarticletitle{ElasticFusion: Real-time dense SLAM and light
  source estimation}.
\newblock \bibinfo{journal}{\emph{The International Journal of Robotics
  Research}} \bibinfo{volume}{35}, \bibinfo{number}{14} (\bibinfo{year}{2016}),
  \bibinfo{pages}{1697--1716}.
\newblock


\bibitem[\protect\citeauthoryear{Wood, Azuma, Aldinger, Curless, Duchamp,
  Salesin, and Stuetzle}{Wood et~al\mbox{.}}{2000}]%
        {Wood2000}
\bibfield{author}{\bibinfo{person}{Daniel~N. Wood}, \bibinfo{person}{Daniel~I.
  Azuma}, \bibinfo{person}{Ken Aldinger}, \bibinfo{person}{Brian Curless},
  \bibinfo{person}{Tom Duchamp}, \bibinfo{person}{David~H. Salesin}, {and}
  \bibinfo{person}{Werner Stuetzle}.} \bibinfo{year}{2000}\natexlab{}.
\newblock \showarticletitle{Surface Light Fields for 3D Photography}. In
  \bibinfo{booktitle}{\emph{Proceedings of the 27th Annual Conference on
  Computer Graphics and Interactive Techniques}}
  \emph{(\bibinfo{series}{SIGGRAPH '00})}. \bibinfo{publisher}{ACM
  Press/Addison-Wesley Publishing Co.}, \bibinfo{address}{New York, NY, USA},
  \bibinfo{pages}{287--296}.
\newblock
\showISBNx{1-58113-208-5}
\urldef\tempurl%
\url{https://doi.org/10.1145/344779.344925}
\showDOI{\tempurl}


\bibitem[\protect\citeauthoryear{Worrall, Garbin, Turmukhambetov, and
  Brostow}{Worrall et~al\mbox{.}}{2017}]%
        {worrall2017interpretable}
\bibfield{author}{\bibinfo{person}{Daniel~E Worrall},
  \bibinfo{person}{Stephan~J Garbin}, \bibinfo{person}{Daniyar Turmukhambetov},
  {and} \bibinfo{person}{Gabriel~J Brostow}.} \bibinfo{year}{2017}\natexlab{}.
\newblock \showarticletitle{Interpretable transformations with encoder-decoder
  networks}. In \bibinfo{booktitle}{\emph{Proc. ICCV}},
  Vol.~\bibinfo{volume}{4}.
\newblock


\bibitem[\protect\citeauthoryear{Yan, Yang, Yumer, Guo, and Lee}{Yan
  et~al\mbox{.}}{2016}]%
        {yan2016perspective}
\bibfield{author}{\bibinfo{person}{Xinchen Yan}, \bibinfo{person}{Jimei Yang},
  \bibinfo{person}{Ersin Yumer}, \bibinfo{person}{Yijie Guo}, {and}
  \bibinfo{person}{Honglak Lee}.} \bibinfo{year}{2016}\natexlab{}.
\newblock \showarticletitle{Perspective transformer nets: Learning single-view
  3d object reconstruction without 3d supervision}. In
  \bibinfo{booktitle}{\emph{Proc. NIPS}}. \bibinfo{pages}{1696--1704}.
\newblock


\bibitem[\protect\citeauthoryear{Zeng, Zhao, Zheng, and Liu}{Zeng
  et~al\mbox{.}}{2013}]%
        {zeng2013octree}
\bibfield{author}{\bibinfo{person}{Ming Zeng}, \bibinfo{person}{Fukai Zhao},
  \bibinfo{person}{Jiaxiang Zheng}, {and} \bibinfo{person}{Xinguo Liu}.}
  \bibinfo{year}{2013}\natexlab{}.
\newblock \showarticletitle{Octree-based fusion for realtime 3D
  reconstruction}.
\newblock \bibinfo{journal}{\emph{Graphical Models}} \bibinfo{volume}{75},
  \bibinfo{number}{3} (\bibinfo{year}{2013}), \bibinfo{pages}{126--136}.
\newblock


\bibitem[\protect\citeauthoryear{Zheng, Colburn, Agarwala, Agrawala, Salesin,
  Curless, and Cohen}{Zheng et~al\mbox{.}}{2009}]%
        {Zheng2009}
\bibfield{author}{\bibinfo{person}{Ke~Colin Zheng}, \bibinfo{person}{Alex
  Colburn}, \bibinfo{person}{Aseem Agarwala}, \bibinfo{person}{Maneesh
  Agrawala}, \bibinfo{person}{David Salesin}, \bibinfo{person}{Brian Curless},
  {and} \bibinfo{person}{Michael~F. Cohen}.} \bibinfo{year}{2009}\natexlab{}.
\newblock \showarticletitle{Parallax photography: creating 3D cinematic effects
  from stills}. In \bibinfo{booktitle}{\emph{Proc. Graphics Interface}}.
  \bibinfo{publisher}{{ACM} Press}, \bibinfo{pages}{111--118}.
\newblock


\bibitem[\protect\citeauthoryear{Zhou and Koltun}{Zhou and Koltun}{2014}]%
        {zhou2014color}
\bibfield{author}{\bibinfo{person}{Qian-Yi Zhou} {and} \bibinfo{person}{Vladlen
  Koltun}.} \bibinfo{year}{2014}\natexlab{}.
\newblock \showarticletitle{Color map optimization for 3D reconstruction with
  consumer depth cameras}.
\newblock \bibinfo{journal}{\emph{ACM Transactions on Graphics (TOG)}}
  \bibinfo{volume}{33}, \bibinfo{number}{4} (\bibinfo{year}{2014}),
  \bibinfo{pages}{155}.
\newblock


\bibitem[\protect\citeauthoryear{Zhou, Tucker, Flynn, Fyffe, and Snavely}{Zhou
  et~al\mbox{.}}{2018}]%
        {Zhou:2018}
\bibfield{author}{\bibinfo{person}{Tinghui Zhou}, \bibinfo{person}{Richard
  Tucker}, \bibinfo{person}{John Flynn}, \bibinfo{person}{Graham Fyffe}, {and}
  \bibinfo{person}{Noah Snavely}.} \bibinfo{year}{2018}\natexlab{}.
\newblock \showarticletitle{Stereo Magnification: Learning View Synthesis Using
  Multiplane Images}.
\newblock \bibinfo{journal}{\emph{ACM Trans. Graph.}} \bibinfo{volume}{37},
  \bibinfo{number}{4}, Article \bibinfo{articleno}{65} (\bibinfo{date}{July}
  \bibinfo{year}{2018}), \bibinfo{numpages}{12}~pages.
\newblock
\showISSN{0730-0301}


\bibitem[\protect\citeauthoryear{Zhou, Tulsiani, Sun, Malik, and Efros}{Zhou
  et~al\mbox{.}}{2016}]%
        {zhou2016view}
\bibfield{author}{\bibinfo{person}{Tinghui Zhou}, \bibinfo{person}{Shubham
  Tulsiani}, \bibinfo{person}{Weilun Sun}, \bibinfo{person}{Jitendra Malik},
  {and} \bibinfo{person}{Alexei~A Efros}.} \bibinfo{year}{2016}\natexlab{}.
\newblock \showarticletitle{View synthesis by appearance flow}. In
  \bibinfo{booktitle}{\emph{Proc. ECCV}}. Springer, \bibinfo{pages}{286--301}.
\newblock


\end{thebibliography}

\end{document}